%% file: main.tex
\documentclass[10pt,twocolumn,letterpaper]{article}

\usepackage[pagenumbers]{cvpr}      % To produce the REVIEW version
\usepackage{makecell}
\usepackage{bm}
\usepackage{float}
\usepackage{afterpage}

\usepackage{multirow}
\usepackage[normalem]{ulem}
\usepackage{setspace}
\usepackage{pifont}
\usepackage{float}
\input{preamble}

\definecolor{cvprblue}{rgb}{0.21,0.49,0.74}
\usepackage[pagebackref,breaklinks,colorlinks,citecolor=cvprblue]{hyperref}

\def\paperID{9757} % *** Enter the Paper ID here
\def\confName{CVPR}
\def\confYear{2024}

\begin{document}
% \title{\LaTeX\ Author Guidelines for \confName~Proceedings}
% \title{Layout2Urban: Layout-Guided Urban Radiance Field Generation}
\title{Urban Architect: Steerable 3D Urban Scene Generation with Layout Prior}
% \title{Urban Architect: Steerable Urban Radiance Field Generation with Layout Prior}
%\title{Urban Architect: Steering Urban Radiance Field Generation with Layout Prior}
% \title{Urban Architect: Steerable 3D Urban Scene Generation with Layout Prior}
% \title{Urban Architect: Steerable 3D Urban Scene Generation via Layout as Prior}

%%%%%%%%% AUTHORS - PLEASE UPDATE
\author{Fan Lu\textsuperscript{1}~~~Kwan-Yee Lin\textsuperscript{2 \dag}~~~Yan Xu\textsuperscript{3}~~~Hongsheng Li\textsuperscript{2,4,5}~~~Guang Chen\textsuperscript{1 \dag}~~~Changjun Jiang\textsuperscript{1}\\
\textsuperscript{1}Tongji University~~~\textsuperscript{2}Shanghai AI Laboratory~~~\textsuperscript{3}University of Michigan\\~~~\textsuperscript{4}The Chinese University of Hong Kong~~~\textsuperscript{5}CPII\\
{\tt\small\{lufan,guangchen,cjjiang\}@tongji.edu.cn} \\
\tt\small{linjunyi9335@gmail.com, yxumich@umich.com, hsli@ee.cuhk.edu.hk}
}

\twocolumn[
{%
\maketitle
\vspace{-1cm}
\begin{figure}[H]
\hsize=\textwidth % 
\centering
\setlength{\abovecaptionskip}{1ex}
 \includegraphics[width=0.95\textwidth]{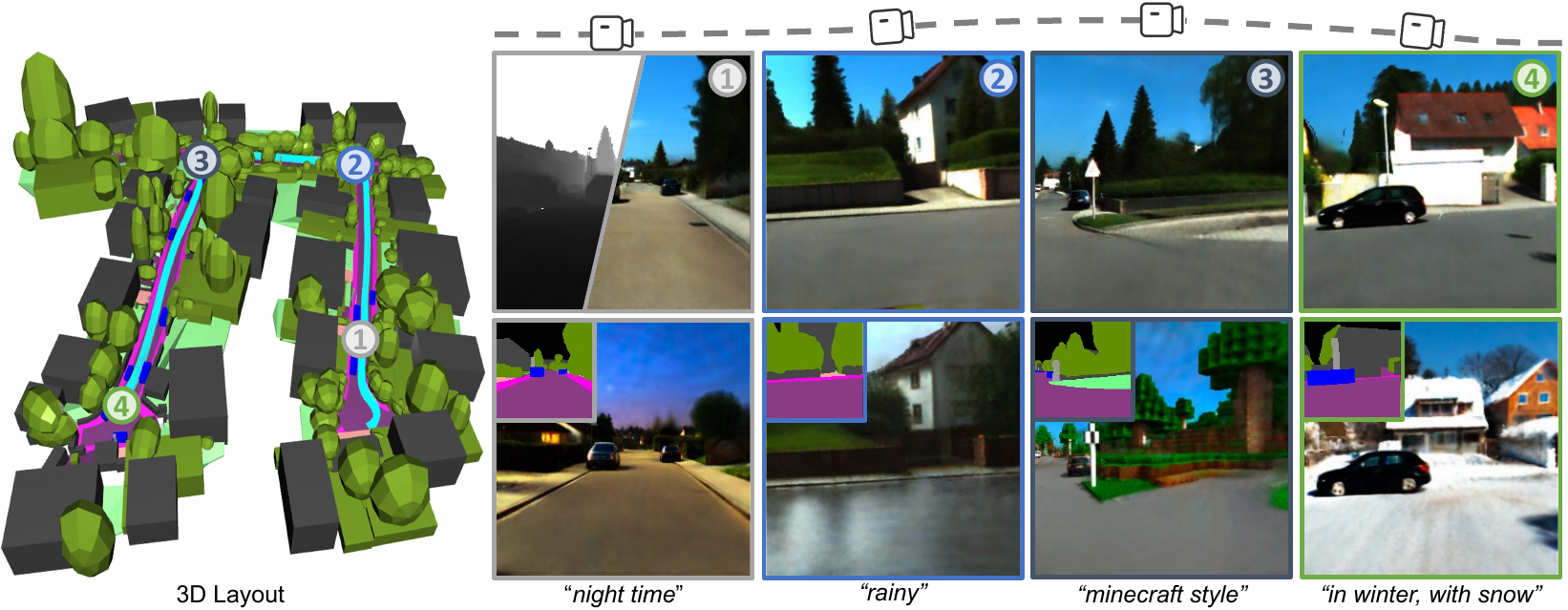}
% \vspace{-1ex}
\caption{
\textbf{Urban Architect for steerable 3D urban scene generation.} We present Urban Architect, a method to generate steerable 3D urban scenes by introducing 3D layout as an additional prior to complement textual descriptions. The framework enjoys three key properties: $(1)$ Large-scale urban scene creation. The scale goes beyond $1000$m driving distance in our experiments. $(2)$ High quality. The generated scene enables photo-realistic rendering (upper row) and obeys geometric consistency (the left side of the first camera's upper image). $(3)$ Steerable creation process. It supports various scene editing effects by fine-tuning the generated scene in a breeze (\textit{e.g.,} style editing (lower row)). Project page: \href{https://urbanarchitect.github.io/}{https://urbanarchitect.github.io}.
}\label{fig:teaser}
\end{figure}
% \vspace{-2ex}
}
]

\let\thefootnote\relax\footnotetext{\dag corresponding authors.}
% }

% \maketitle
\input{sec/0_abstract}    
\input{sec/1_intro}

\input{sec/2_relatedwork}
\input{sec/3_method}
\input{sec/4_experiments}
\input{sec/5_conclusion}
% {
%     \small
%     \bibliographystyle{ieeenat_fullname}
%     \bibliography{main}

{
    \small
    \bibliographystyle{ieeenat_fullname}
    \bibliography{main}
}

\end{document}

%% file: preamble.tex
\usepackage[dvipsnames]{xcolor}

%% file: sec/0_abstract.tex
\begin{abstract}

%%% ---- simplicated version -----
Text-to-3D generation has achieved remarkable success in digital object creation, attributed to the utilization of large-scale text-to-image diffusion models. Nevertheless, there is no paradigm for scaling up the methodology to urban scale. The complexity and vast scale of urban scenes, characterized by numerous elements and intricate arrangement relationships, present a formidable barrier to the interpretability of ambiguous textual descriptions for effective model optimization.
In this work, we surmount the limitations via a paradigm shift in the current text-to-3D methodology, accomplished through a compositional 3D layout representation serving as an additional prior. The 3D layout comprises a set of semantic primitives with simple geometric structures (\textit{e.g.,} cuboids, ellipsoids, and planes), and explicit arrangement relationships. It complements textual descriptions, and meanwhile enables steerable generation. Based on the 3D layout representation, we propose two modifications to the current text-to-3D paradigm -- $(1)$ We introduce Layout-Guided Variational Score Distillation (LG-VSD) to address model optimization inadequacies. It incorporates the geometric and semantic constraints of the 3D layout into the the fabric of score distillation sampling process, effectuated through an elegant formula extension into a conditional manner. $(2)$ To handle the unbounded nature of the urban scenes, we represent the 3D scene with a Scalable Hash Grid structure, which incrementally adapts to the growing scale of urban scenes. Extensive experiments substantiate the robustness of our framework, showcasing its capability to scale text-to-3D generation to large-scale urban scenes that cover over $1000$m driving distance for the first time. We also present various scene editing demonstrations (\textit{e.g.,} style editing, object manipulation, etc.), showing the complementary powers of both 3D layout prior and text-to-image diffusion models in our framework for steerable urban scene generation.

\end{abstract}

%% file: sec/1_intro.tex
\section{Introduction}
\label{sec:intro}
A steerable paradigm to create 3D urban scenes with realism and flexibility hosts utmost significance to various applications (\textit{e.g.,} autonomous driving simulation, virtual reality, games, and \etc.). In previous research, tasks related to urban-scale scene creation, such as 3D-aware image synthesis~\cite{bahmani2023cc3d,yang2023urbangiraffe,lin2023infinicity,xie2023citydreamer} and 3D scene reconstruction~\cite{lu2023urban,liu2023real}, have made strides with different methodologies and specializations. However, they are all stuck in the dilemma of trading-off among scene scale, quality, flexibility, and geometric consistency. This prompts a pivotal inquiry: {\textit{What methodology could facilitate the realization of steerable 3D urban scene creation?}}

Text-to-3D generation offers a promising direction. Rooted in the flexible and versatile textual condition, it enjoys the excellence of both remarkable quality from text-to-image diffusion models, and the geometric consistency from 3D representations.  
Notable examples include, among many others, DreamFusion~\cite{poole2023dreamfusion}, Magic3D~\cite{lin2023magic3d} and ProlificDreamer~\cite{wang2023prolificdreamer}, \etc. At the heart of these advances is the idea of Score Distillation Sampling (SDS)~\cite{poole2023dreamfusion}. It brings the best of two worlds together by optimizing a 3D model via aligning the distribution of rendered images with the target distribution derived from the diffusion model. Nevertheless, the ascendancy faces a formidable challenge when directly extending the paradigm to urban-scale scenes.
% text-to-image 

% (\eg, neural radiance field (NeRF)~\cite{mildenhall2021nerf}, tetrahedron mesh~\cite{lin2023magic3d}, \etc) 

% Text prompt, due to its flexibility and versatility, has become one of the most common signals for 3D content creation. Recently, large-scale text-to-image diffusion models have significantly advanced text-to-3D generation. The core of most current text-to-3D methods is Score Distillation Sampling (SDS). SDS optimizes a 3D model (\eg, neural radiance field (NeRF)~\cite{mildenhall2021nerf}, tetrahedron mesh~\cite{lin2023magic3d}, \etc) by aligning the distribution of rendered images with the target distribution derived from the pre-trained diffusion model under the textual condition. Benefiting from the capacity of large-scale diffusion models, SDS-based methods are capable of generating high-definite 3D content.

% Text prompt, due to its flexibility and versatility, has become one of the most common signals for 3D content creation. Recently, large-scale text-to-image diffusion models have significantly advanced text-to-3D generation. The core of most current text-to-3D methods is Score Distillation Sampling (SDS). SDS optimizes a 3D model (\eg, neural radiance field (NeRF)~\cite{mildenhall2021nerf}, tetrahedron mesh~\cite{lin2023magic3d}, \etc) by aligning the distribution of rendered images with the target distribution derived from the pre-trained diffusion model under the textual condition. Benefiting from the capacity of large-scale diffusion models, SDS-based methods are capable of generating high-definite 3D content.

Urban scenes, characterized by high complexity and unboundedness, pose two key challenges for existing text-to-3D paradigms to create high-quality 3D content: $(1)$ {\textit{How to optimize a scene with dense concepts and intricate arrangement relationships? }}Text prompts are inherently ambiguous and struggle to provide fine-grained conditional signals. This granularity gap between text and real-world urban scene content triggers the insufficient optimization predicament of SDS-based methodology, where SDS (and its variant VSD~\cite{wang2023prolificdreamer}) would fail to capture the complex multimodal distribution of real-world urban scene, given only the ambiguous textual condition. $(2)$ {\textit{How to represent a 3D scene that is unbounded and vast in spatial scale?}} Previous text-to-3D methods mainly utilize bounded models such as bounded NeRF~\cite{mildenhall2021nerf,poole2023dreamfusion} and tetrahedron mesh~\cite{shen2021deep,lin2023magic3d} as 3D representation, which is inapplicable to unbounded urban scenes with arbitrary scales.

In this work, we present Urban Architect, a lightweight solution that utilizes the power of a compositional 3D layout representation to overcome the above dilemmas and extend the current text-to-3D paradigm to urban scale. The 3D layout representation is an in-hand specific prior, which could complement the ambiguities of text prompts in a breeze. As shown in Fig.~\ref{fig:teaser}, the 3D layout comprises a set of semantic primitives, where instances of an urban scene are represented using these simple geometric structures (\eg, cuboids for buildings and cars, ellipsoids for vegetation, and planes for roads), making it effective for guiding model optimization and easy for users to assemble. Centered on the representation, we provide two modifications to the current text-to-3D paradigm.

% representation possesses several superior properties: \textit{(1) User-friendly construction.} 3D layout 

% consists of a set of simple semantic primitives, where instances of an urban scene are represented using these simple geometric structures (\eg, {\color{red}{cuboids for building and cars, ellipsoid for vegetation, plane for road)}, making it easy for users to assemble. \textit{(2) Effective guidance.} The 3D layout provides both geometric and semantic constraints of the scene distribution by nature, which could offer robust guidance for model optimization. \textit{(3) Fine-grained control and editing} The compositional nature of the layout representation enables instance-level control and editing of the scene.} Centered on the representation, we provide two modifications to the current text-to-3D paradigm. 
% % \textit{(3) Fine-grained control and editing.} The compositional nature of the representation enables instance-level control and editing of the scene. 

To address model optimization inadequacies, we propose Layout-Guided Variational Score Distillation (LG-VSD). It is a conditional extension of SDS and VSD, which integrates additional conditions from the 3D layout into the optimization process. Concretely, we first map the 3D layout constraints to 2D space by rendering 2D semantic and depth maps at arbitrary camera views, which results in a sequence of 2D conditions with 3D consistency. To preserve the original power of the pre-trained 2D diffusion model, and meanwhile inject our explicit constraints, we leverage ControlNet~\cite{zhang2023adding} to integrate conditions from the rendered semantic and depth maps. Consequently, we could control the score distilling process with the 2D conditional signals derived from the 3D layout. As these signals 
provide explicit prior to semantic and geometric distribution of the scene, they could enforce the optimization process to match the 3D layout consistency and therefore lead to high-fidelity results. Moreover, as LG-VSD only changes the sampling process, the modification is seamlessly grafted to the original diffusion model. Thus, we could harness the inherent refinement ability of the diffusion model to further improve the rendering quality. To ensure the semantic and geometric consistency between coarse and refined stages, we introduce a layout-aware refinement strategy to fine-tune the generated scene in practice.
%Concretely, {\color{red}to reduce the dimension, we first map} the 3D layout constraints to 2D image space by rendering 2D semantic and depth maps at arbitrary camera views. {\color{red} In this way, we obtain a sequence of 2D conditions that are 3D consistent in semantic and geometric level.} {\color{red} To preserve the original power of the pre-trained 2D diffusion model, and meanwhile inject our explicit constraints}, we leverage ControlNet~\cite{zhang2023adding} to integrate conditions from the rendered semantic and depth maps{\color{red}. With such layout-based conditions, we could turn the process of score sampling into a conditional manner, where the 2D conditional signals control the distilling process of VSD. The control in 2D space constrains the semantic and geometric distribution of the scene to match the 3D layout prior, leading to layout-consistent and high-fidelity generation results. Since the LG-VSD only changes the sampling process, the modification is seamlessly grafted to XXX. Thus, we could also harness the inherent refinement ability of the pre-trained diffusion model to further improve the rendering quality. This could be achieved by simply adding a layout-aware refinement strategy to fine-tune the generated scene.

We further present a Scalable Hash Grid (SHG) structure as a pragmatical and effective means to represent 3D unbounded urban scenes. Uniquely tailored for urban scenarios, SHG embodies two key design principles. Firstly, instead of encapsulating the entire scene within a single 3D model, SHG is a hash structure that is adaptively updated with the dynamic change of camera trajectories. This incremental functionality liberates it from a fixed spatial range that current 3D representations suffer. Second, we exploit the geometric information of the 3D layout to constrain the sampling space of the hash grid, thereby enhancing the overall rendering quality.

% To further improve the rendering quality, we harness the inherent refinement ability of the pre-trained diffusion model and design a layout-aware refinement strategy to finetune the generated radiance field.
% % Given the inherent ambiguity of textual prompts, the target distribution from the diffusion model is diverse, leading to noisy guidance for model optimization. 

Extensive experiments substantiate the capacity of the proposed method to generate complex 3D urban scenes with high quality, diversity, and layout consistency based on the provided 3D layouts. The robust constraints from the 3D layout prior, combined with the scalable hash grid, facilitate the generation of large-scale urban scenes covering over $1000$m driving distance. By leveraging the flexible and fine-grained constraints of our layout representation, along with the text-based style control, our pipeline supports various scene editing effects, including instance-level editing (\eg, object manipulation) and style editing (\textit{e.g.}, transferring city style and weather).

%% file: sec/2_relatedwork.tex
\section{Related Work}
\label{sec:relatedwork}

%{\ky{(2.1) 3D representations for Urban Scene. (1) why choose nrf? (2) the drawback of nrf\\(2.2) Text-to-image. diversity, fidelity\\(2.3) Text-to-3D. (1)the core element to success is sds. (2) drawbacks such as over-smooth, (3) cannot scale up to urban scene.\\(2.4 3D-aware image synthesis.*struggle to generate high-quality large-scale urban scenes* is ambiguous. (1) geometry consistency problem, (2) overfitting )}

\noindent \textbf{3D Representations for Urban Scene.} Various 3D representations have been proposed to model 3D scenes. Explicit representations, such as point clouds and meshes, pose challenges for optimization through purely 2D supervisions given their discrete nature, especially in large-scale scenes due to the high complexity and ill-posed optimization process. NeRF~\cite{mildenhall2021nerf} adopts implicit neural representations to encode densities and colors of the scene and exploits volumetric rendering for view synthesis, which can be effectively optimized from 2D multi-view images. Given its superior performance in view synthesis, NeRF has been used to model diverse 3D scenes. Numerous works have enhanced NeRF in terms of rendering quality~\cite{barron2021mip,barron2022mip,barron2023zipnerf,hu2023tri}, efficiency~\cite{sun2022direct,chen2022tensorf,muller2022instant,fridovich2022plenoxels,kerbl20233d}, \emph{etc}. Some works have also extended NeRF to large-scale urban scenes~\cite{tancik2022block,lu2023urban,rematas2022urban,xu2023grid,xiangli2022bungeenerf}. For example,  URF~\cite{rematas2022urban} adopts LiDAR point clouds to facilitate geometric learning and uses a separate network to model the sky. However, prior works often assume a fixed scene range and contract distant structures, encountering challenges when extending to urban scenes with arbitrary scales.

\noindent \textbf{Text-to-3D.} 
Early text-to-3D methods adopted CLIP~\cite{radford2021learning} to guide the optimization but encountered difficulties in creating high-quality 3D content~\cite{jain2022zero,sanghi2022clip,mohammad2022clip,hong2022avatarclip}. Recently, large-scale diffusion models~\cite{rombach2022high,balaji2022ediffi,saharia2022photorealistic} have demonstrated significant success in text-to-image generation, capable of generating 2D images with high fidelity and diversity. They further enable fine-grained control and editing by employing additional conditioning models~\cite{zhang2023adding,meng2021sdedit}. For example, ControlNet~\cite{zhang2023adding} incorporates 2D pixel-aligned conditional signals to control the generation. SDEdit~\cite{meng2021sdedit} achieves conditional image synthesis and editing using reverse SDE. The achievements in 2D content creation have propelled text-to-3D generation~\cite{poole2023dreamfusion,lin2023magic3d,wang2023prolificdreamer,Chen_2023_ICCV,metzer2023latent,huang2023dreamtime,tang2023dreamgaussian}.
As the core of most current text-to-3D methods, Score Distillation Sampling (SDS)~\cite{poole2023dreamfusion} optimizes a 3D model by aligning 2D images rendered at arbitrary viewpoints with the distribution derived from a text-conditioned diffusion model. However, SDS suffers from issues such as over-smoothness and over-saturation. VSD~\cite{wang2023prolificdreamer} proposes a particle-based variational framework to enhance the generation quality of SDS. Some method also attempts to incorporate 3D information into the generation process to enhance 3D consistency~\cite{metzer2023latent,chen2023fantasia,jiang2023avatarcraft}. For example, AvatarCraft~\cite{jiang2023avatarcraft} utilize SMPL~\cite{smpl} mesh for 3D human avatar generation and animation. Nonetheless, most prior works are tailored for single object-level generation. Some methods have attempted to achieve scene-level generation, while the scene scales are quite limited~\cite{lin2023componerf, cohen2023set}. When applied to large-scale urban scenes, current text-to-3D methods face challenges in modeling the intricate distribution from the purely text-conditioned diffusion model. Alternative methods have explored incrementally reconstructing and inpainting the scene to create room-scale~\cite{Hollein_2023_ICCV} or zoom-out~\cite{fridman2023scenescape} scenes. However, this paradigm is susceptible to cumulative errors and poses challenges in scalability for larger scenes.

\noindent \textbf{3D Generative Models.} Many 3D generative models leverage GAN~\cite{goodfellow2014generative}, VAE~\cite{kingma2013auto}, and diffusion model~\cite{ho2020denoising} to model the distribution of 3D representations~\cite{wu2016learning,achlioptas2018learning,zeng2022lion,wang2023rodin}. However, this paradigm requires a large amount of 3D data and is often limited to single-object generation. NeuralField-LDM~\cite{kim2023neuralfield} trains a scene-level 3D diffusion model by preparing abundant voxel-based neural fields, while the scales are limited to the pre-defined 3D voxel grid. Another line of work employs 3D-aware GANs to optimize a 3D model from 2D image collections. Early 3D-aware GANs are mainly designed for object-level scene generation~\cite{deng2022gram,schwarz2020graf,chan2022efficient,gao2022get3d}. Recently, some methods have extended the pipeline to larger scenes including indoor scenes~\cite{bautista2022gaudi,devries2021unconstrained}, landscapes~\cite{chen2023scene}, and urban scenes~\cite{bahmani2023cc3d,yang2023urbangiraffe,lin2023infinicity}). For example, GSN~\cite{devries2021unconstrained} proposes a local latent grid representation for indoor scene generation. SceneDreamer~\cite{chen2023scene} aims at generating landscapes, which contain repeated textures and structures. CityDreamer~\cite{xie2023citydreamer} further enhances SceneDreamer's foreground generation quality for city scene generation. The key modification is separating building instances from other background objects. The main insight of these two methods to achieve unbounded scene generation is splitting scenes into local windows, and encoding local scene features implicitly in hash grids. However, their procedural generation process constrains the diversity and the handling of structural details, which are critical in urban scene generation. Besides, the implicit scene encoding-based hash representation limits the ability to handle complex scenes. Some methods have also attempted to incorporate user inputs to control urban scene generation. CC3D~\cite{bahmani2023cc3d} integrates 2D layout conditions in the 3D generative model for urban scene generation. UrbanGIRAFFE~\cite{yang2023urbangiraffe} and InfiniCity~\cite{lin2023infinicity} incorporate 3D voxel grids to enhance quality and scalability. Nevertheless, owing to inherent constraints such as limited training data and model capacity, they still suffer from structural distortions and the lack of details, and often overfit to the training scenarios.

%% file: sec/3_method.tex
\begin{figure*}[htbp]
\hsize=\textwidth % 
\centering
\includegraphics[width=1.\textwidth]{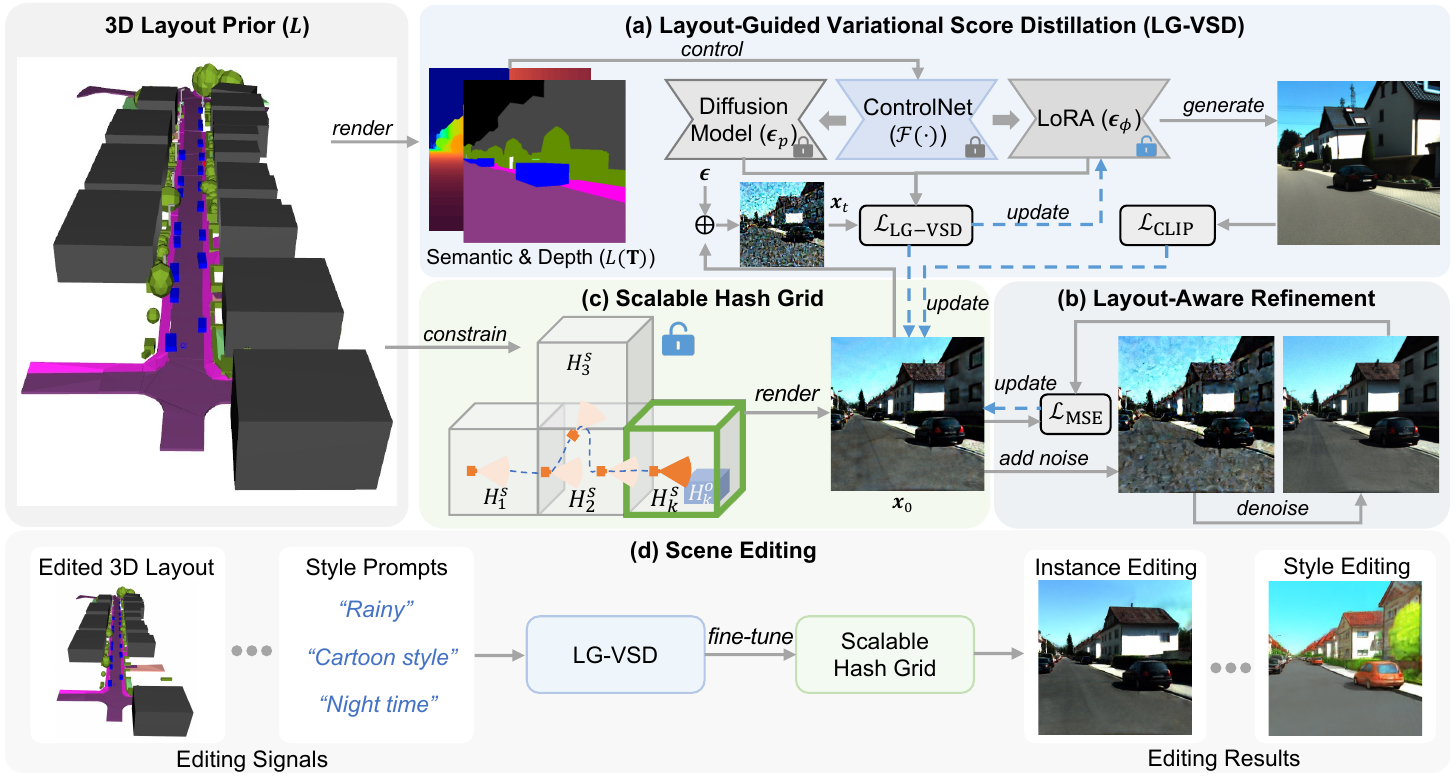}
\caption{\textbf{Overview of Urban Architect.} We introduce Urban Architect, a method that generates urban-scale 3D scenes with 3D layout instruction and textural descriptions. The scene is represented by a neural field that is optimized by distilling a pre-trained diffusion model in a conditional manner. (a) Rather than relying solely on the text-based guidance, we propose to control the distilling process of Variational Score Distillation (VSD) via the 3D layout of the desired scene, introducing Layout-Guided Variational Score Distillation (LG-VSD). (b) We refine the local details via a layout-aware refinement strategy. (c) To model unbounded urban scenes, we discretize the 3D representation into a scalable hash grid. (d) We support various scene editing effects by fine-tuning the generated scene.
}\label{fig:framework}
% \vspace{-4ex}
\end{figure*}
\section{Method}
We propose Urban Architect for large-scale 3D urban scene generation. The overall pipeline is illustrated in Fig.~\ref{fig:framework}. Given the 3D layout of the desired scene, we first optimize the scene by distilling a pre-trained text-to-image diffusion model via the proposed Layout-Guided Variational Score Distillation (LG-VSD), which complements textual descriptions with additional 3D layout constraints (Sec.~\ref{sec:lgvsd}). 
Then, the generated scene is refined by a layout-aware refinement strategy to enhance quality (Sec.~\ref{sec:refine}). The 3D scene is represented using a scalable hash grid representation to tackle the unbounded nature of urban scenarios (Sec.~\ref{sec:hash}). Finally, we demonstrate the ability of our pipeline to achieve diverse scene editing effects by fine-tuning the generated scene (Sec.~\ref{sec:edit}).

\subsection{Preliminaries}
\noindent \textbf{Score Distillation Sampling (SDS).} SDS achieves text-to-3D generation by distilling a pre-trained text-to-image diffusion model $\bm{\epsilon}_p$ to optimize a differentiable 3D representation parameterized by $\theta$. SDS encourages the image $\bm{x}_0=g(\theta,\mathbf{T})$ rendered at an arbitrary camera view $\mathbf{T}$ to match the distribution $p_0(\bm{x}_0|y)$ derived from the diffusion model conditioned on the text prompt $y$. In practice, SDS optimizes the 3D model by adding random noise $\bm{\epsilon}\sim \mathcal{N}(\bm{0},\bm{I})$ to $\bm{x}_0$ and calculating the difference between the predicted noise from the diffusion model $\bm{\epsilon}_p$ and the added noise $\bm{\epsilon}$. The gradient of SDS loss can be formulated as:
\begin{small}
\begin{equation}
\begin{split}
    \nabla_{\theta}\mathcal{L}_{\rm SDS}(\theta)\triangleq \mathbb{E}_{t,\bm{\epsilon},\mathbf{T}}\bigg[\omega (t)(\bm{\epsilon}_{p}(\bm{x}_t, t, y)-
    \bm{\epsilon})\frac{\partial \bm{g}(\theta, \mathbf{T})}{\partial \theta}\bigg],
\end{split}
\end{equation}
\end{small}
where $t\sim \mathcal{U}(0.02, 0.98)$, $\omega(t)$ weights the loss given the time step $t$, and $\bm{x}_t$ is the randomly perturbed rendered image. Although SDS can produce reasonable 3D content aligned with the text prompt, it exhibits issues such as over-saturation, over-smoothing, and low diversity.

\noindent \textbf{Variational Score Distillation (VSD).} To address the limitations of SDS, instead of optimizing a single sample point, VSD proposes to optimize a distribution $q_0^{\mu}(\bm{x}_0|y)$ of possible 3D representations with $\mu (\theta|y)$ corresponding to the text prompt $y$. VSD adopts a set of 3D parameters $\{\theta\}_{i=1}^n$ as particles to represent the scene distribution and optimizes the 3D model by matching the score of noisy real images and that of noisy rendered images at each time step $t$. In practice, the score function of noisy rendered images is estimated by optimizing a low-rank adaptation (LoRA) model $\bm{\epsilon}_{\phi}$~\cite{hu2021lora} of the pre-trained diffusion model $\bm{\epsilon}_p$. The gradient of VSD loss can be formulated as:
\begin{small}
\begin{equation}
\begin{split}
    \nabla_{\theta}\mathcal{L}_{\rm VSD}(\theta)\triangleq \mathbb{E}_{t,\bm{\epsilon},\mathbf{T}}\bigg[\omega (t)(\bm{\epsilon}_{p}(\bm{x}_t, t, y)- \\
    \bm{\epsilon}_{\phi}(\bm{x}_t, t, \mathbf{T}, y))\frac{\partial \bm{g}(\theta, \mathbf{T})}{\partial \theta}\bigg].
\end{split}
\end{equation}
\end{small}
VSD proves to be effective in alleviating issues of SDS such as over-saturation and over-smoothing, and successfully generates high-fidelity and diverse 3D content. 
%We use a simple version of VSD with a single particle and observe better performance than SDS.}
% to the target distribution $p_0(\bm{x}_0|y)$.

\subsection{Layout-Guided Variational Score Distillation}\label{sec:lgvsd}

VSD relies on text prompts to guide model optimization. However, given the inherent ambiguity of text prompts $y$, it is challenging for VSD to estimate the distribution of a complex urban scene from a diverse target distribution $p_0(\bm{x}_0|y)$. Therefore, introducing additional constraints to obtain a more compact distribution is crucial for high-quality 3D urban scene generation. A straightforward approach is to simply fine-tune the diffusion model to constrain the distribution to the desired urban scene. However, this operation can only provide style constraints and VSD still struggles to capture the intricate geometric and semantic distribution of urban scenes, leading to unsatisfactory results.

% vanilla VSD: try to find a trivial solution\\
% VSD + finetuned: no trivial solution, much more details, thus messy\\
% LG-VSD: more constraint on the semantic and geometric distribution
% To enable the generation process of VSD to perceive the 3D scene layout, it is crucial to introduce layout guidance in the pre-trained diffusion model. 

Given the above observations, we introduce 3D layout $L$ as a prior to provide additional constraints to the target distribution with a formula shift (\ie, $p_0(\bm{x}_0|y)\rightarrow p_0(\bm{x}_0|y,L)$). Nonetheless, the direct integration of the 3D information into the 2D framework poses a challenge. We observe that the projected semantic and depth maps provide a comprehensive description of the 3D layout in 2D space. They offer both semantic and geometric constraints of the scene, and inherently exhibit multi-view consistency. Thus, we propose to condition the distilling process via the 2D semantic and depth maps rendered at the given camera pose $\mathbf{T}$. Specifically, we first train a ControlNet~\cite{zhang2023adding} $\mathcal{F}$ by utilizing the 2D semantic and depth maps rendered from 3D layouts as conditions, along with the corresponding 2D images as ground truths. 
% (\ie, $\mu(\theta|y,L)$, where $L$ is the 3D layout)
%we train a ControlNet $\mathcal{F}$ using the 2D semantic and depth maps rendered from 3D layout as conditions and the corresponding 2D images as ground truths. 
% We start by projecting the 3D information in the 3D prior onto 2D space by rendering 2D semantic and depth maps from arbitrary camera views. Then, we leverage ControlNet~\cite{zhang2023adding} to add additional semantic and depth conditions.
%We transform the 3D We leverage ControlNet~\cite{zhang2023adding} to add additional semantic and depth control. Specifically, we train a ControlNet $\mathcal{F}$ using the 2D semantic and depth maps rendered from 3D layout as conditions and the corresponding 2D images as ground truths. 
At each training step of VSD, we render the 2D signals from the conditional 3D layout $L$ at a randomly sampled camera view $\mathbf{T}$. Subsequently, the features produced by $\mathcal{F}$ are integrated into both the diffusion model $\bm{\epsilon}_p$ and the LoRA model $\bm{\epsilon}_{\phi}$, leading to a compact, scene-specific target distribution $p_0(\bm{x}_0|y,\mathcal{F}(L(\mathbf{T})))$. Formally, the gradient of our layout-guided VSD (LG-VSD) loss can be written as:
\begin{small}
\begin{equation}
\begin{split}
    \nabla_{\theta}\mathcal{L}_{\rm LG\text{-}VSD}(\theta)\triangleq \mathbb{E}_{t,\bm{\epsilon},\mathbf{T}}\bigg[\omega (t)(\bm{\epsilon}_{p}(\bm{x}_t, t, y, \mathcal{F}(L(\mathbf{T}))- \\
    \bm{\epsilon}_{\phi}(\bm{x}_t, t, \mathbf{T}, y, \mathcal{F}(L(\mathbf{T}))))\frac{\partial \bm{g}(\theta, \mathbf{T})}{\partial \theta}\bigg].
\end{split}
\end{equation}
\end{small}

By maximizing the likelihood of the rendered 2D images in the layout-conditioned diffusion model at arbitrary camera views, LG-VSD gradually optimizes the 3D representation $\theta$ to align with the semantic and geometric distribution of the desired scene layout.

Except for the LG-VSD loss, we also use an additional CLIP~\cite{radford2021learning} loss (\ie, $\mathcal{L}_{\rm CLIP}$) to guide the generation. Concretely, we encourage the rendered image $\mathbf{I}_r$ to include consistent content information with the generated image $\mathbf{I}_g$ from the layout-controlled diffusion model. We use the pre-trained image encoder of CLIP to extract features from $\mathbf{I}_r$ and $\mathbf{I}_g$ and calculate the squared L$2$ distance as the loss.

\subsection{Layout-Aware Refinement}\label{sec:refine}
Inspired by SDEdit~\cite{meng2021sdedit}, we observe that the resampling process of diffusion models can effectively refine the rendered images. Specifically, given a random camera pose, the rendered image $\mathbf{I}_r$ is perturbed to $\mathbf{I}_p$ with random noise given the time step $t$. Then $\mathbf{I}_p$ will be denoised to the refined image $\mathbf{I}_f$ using the diffusion model. When $t\rightarrow 0$, the resampling process tends to refine local details of the image $\mathbf{I}_r$ while maintaining its overall structure. Still and all, the geometric and semantic consistencies in long-horizon trajectory cannot be guaranteed, as such a resampling process is agnostic to contextual 3D structures among frames. To this end, we further enable layout-aware refinement by conditioning the denoising steps using the rendered semantic and depth maps, leading to better consistency. The hash grids are then updated with the MSE loss between $\mathbf{I}_r$ and $\mathbf{I}_f$, \emph{i.e.}, $\mathcal{L}_{\rm MSE}=||\mathbf{I}_r-\mathbf{I}_f||_2^2$.

\subsection{Scalable Hash Grid}\label{sec:hash}
\begin{figure}[tb] % 
\centering
\includegraphics[width=1\linewidth]{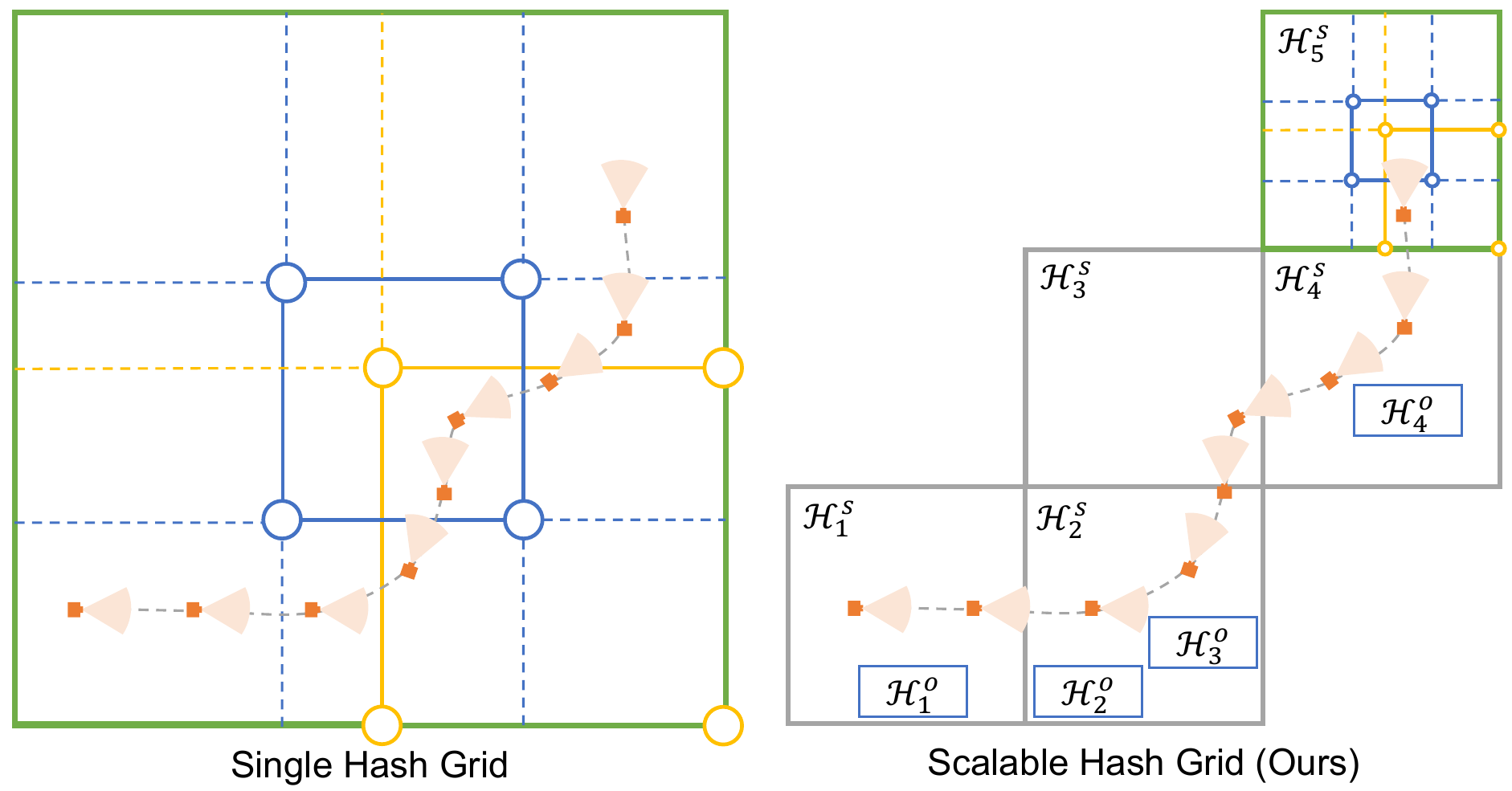}
\caption{\textbf{Illustration of the scalable hash grid representation.} We decomposed the scene into 
a set of stuff and object hash grids (\emph{i.e.}, $\{\mathcal{H}_k^s, \mathcal{H}_k^o\}$). The grids grow with the camera trajectory in a dynamic manner.}\label{fig:hashgrid}
\end{figure}
We introduce a Scalable Hash Grid representation to support unbounded urban scene generation with arbitrary scales. Our representation is based on Zip-NeRF~\cite{barron2023zipnerf}, which combines the fast hash grid-based representation of Instant-NGP~\cite{muller2022instant} with the anti-aliasing ability of Mip-NeRF 360~\cite{barron2022mip}, achieving high-quality and efficient neural rendering. As shown in Fig.~\ref{fig:hashgrid}, instead of modeling the entire scene using a single 3D model, we propose to decompose the scene into a set of stuff grids $\{\mathcal{H}^s_1, \cdots, \mathcal{H}^s_N\}$ and object grids $\{\mathcal{H}^o_1, \cdots, \mathcal{H}^o_M\}$ to enable flexible spatial expansion and further object manipulation.

\noindent\textbf{Stuff Grid.} Each stuff grid $\mathcal{H}^s_k$ models a static space within an axis-aligned bounding box (AABB) centered at $\mathbf{t}_k \in \mathbb{R}^3$ and of size $\mathbf{s}_k\in \mathbb{R}^3$. All structures, excluding objects (\ie, cars in our implementation) are modeled within stuff grids. Given an incoming camera pose $\mathbf{T}$, we first sample points within the camera frustum. Subsequently, the corresponding $\mathcal{H}^s_k$ can be easily determined by transforming the sampled points from world space to the canonical space of each stuff grid. A new stuff grid (\eg, $\mathcal{H}^s_{N+1}$) will be added to the stuff sets when the points fall out of all existing stuff grids. In this way, the stuff representation can be freely updated according to the camera trajectories, making it fully scalable to arbitrary scales.

\noindent\textbf{Object Grid.} Similarly, each object grid $\mathcal{H}_k^o$ also models a space within a 3D bounding box while is parameterized by a rotation matrix $\mathbf{R}_k \in SO(3)$, a translation vector $\mathbf{t}_k\in \mathbb{R}^3$, and a size $\mathbf{s}_k \in \mathbb{R}^3$, as it is not axis-aligned. 

\noindent\textbf{Layout-Constrained Rendering.} For each pixel of the image, we sample a set of points on the camera ray $\mathbf{r}$ and assign each point $\mathbf{x}_i$ to the corresponding stuff or object grid. Thanks to the 3D layout prior, we can simply constrain the sampling space to the interiors of layout instances, which accelerates convergence and leads to better rendering quality. Thereafter, we transform $\mathbf{x}_i$ from world space to the canonical space of the corresponding hash grid (stuff or object) to predict the density $\sigma_i$ and color $c_i$ for subsequent volumetric rendering. Besides, to model the sky region of urban scenes, we follow URF~\cite{rematas2022urban} to predict the color in sky region using the ray direction $\mathbf{d}$ with a separate MLP $\mathcal{R}_{sky}(\cdot)$. Formally, the rendered pixel color $C(\mathbf{r})$ can be calculated as:
\begin{equation}
\begin{split}
    C(\mathbf{r})&=\sum_{i}^{N}T_i\alpha_i c_i + (1-\sum_{i}^{N}T_i\alpha_i)\mathcal{R}_{sky}(\mathbf{d}),\\
    \alpha_i &= 1 - e^{(-\sigma_i\delta_i)},\quad T_i=\prod_{j=1}^{i-1}(1-\alpha_j).
\end{split}
\end{equation}

With the scalable hash grid representation, our pipeline can achieve large-scale scene generation in a breeze. In contrast to the generative hash grid representation employed in SceneDreamer~\cite{chen2023scene}, which encodes scene features implicitly, we explicitly split scenes into stuff and object grids for better scalability.

\subsection{Scene Editing}\label{sec:edit}

\noindent \textbf{Instance-Level Editing.} The compositional 3D layout representation naturally supports instance-level scene editing. For stuff elements (\eg, buildings, trees, \etc), we can delete or insert instances in the scene layout and then fine-tune the generated scene using our LG-VSD for possible missing region completion. For objects (\ie, cars), the manipulation (\eg, rotation, translation, \etc) can be simply applied simultaneously on the instance layout and also the object grid parameters (\ie, $\mathbf{R}_k, \mathbf{t}_k, \mathbf{s}_k$).

\noindent \textbf{Style Editing.} Benefiting from the inherent capacity of the large-scale diffusion model, we can easily transfer the generated scene to various styles. Specifically, given the generated scene, we simply fine-tune it via our LG-VSD, while adding a style text prompt in the distilling process, \eg, \textit{``Cartoon style"}, \textit{``Night time"}, \etc.

\subsection{Layout Construction}
\begin{figure}[tb] % 
\centering
\includegraphics[width=1\linewidth]{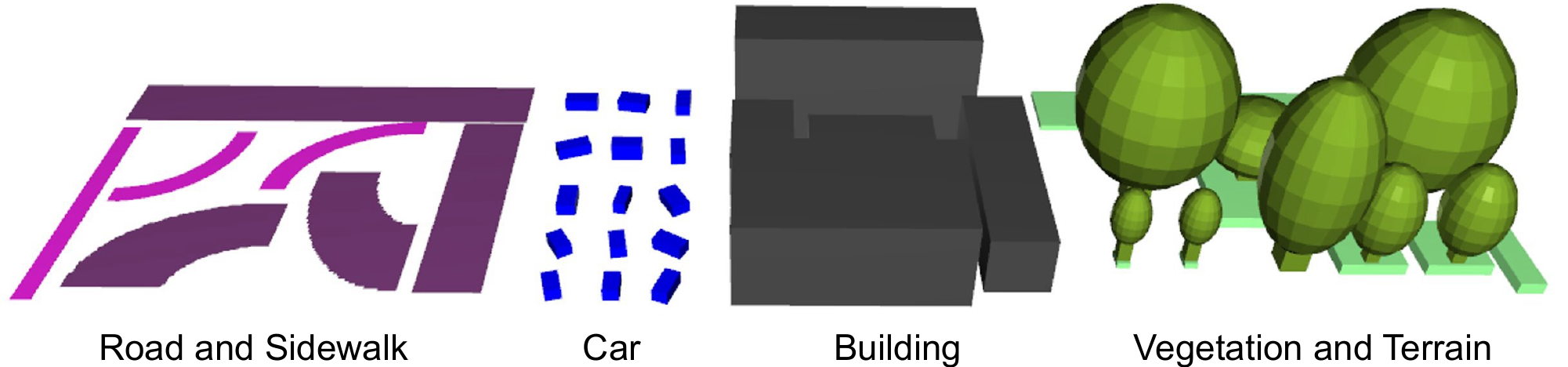}
\caption{\textbf{Basic primitives.} We provide several basic primitives of common objects in urban scenes (\emph{e.g.}, road and sidewalk, car, building, \emph{etc.})}\label{fig:primitives}
\end{figure}

3D layout construction is not the main focus of this work. In practice, we use in-hand layout data provided by the KITTI-360 dataset to train the ControlNet. Nonetheless, there are also several alternatives. For example, users can easily create the desired layout in 3D modeling software (\emph{e.g.}, Blender) given the simple geometric structures of basic primitives. As shown in Fig.~
\ref{fig:primitives}, we provide several basic primitives in urban scenes to ease the process. 

We further provide an alternative method to generate 3D scene layout automatically based on SinDDM~\cite{kulikov2023sinddm}. SinDDM learns the intrinsic distribution of the training image via a multi-scale diffusion process, enabling the generation of new images with arbitrary scales. Specifically, we first compress the 3D layout to the 2D ground plane and obtain a 2D representation of the 3D scene. Then, we train the SinDDM model based on the single 2D example. As shown in Fig.~\ref{fig:sinddm}, the trained model can produce diverse new samples with arbitrary scales and reasonable arrangements. We also generate scenes based on the generated 3D layouts and the rendered results are provided in the bottom row of Fig.~\ref{fig:sinddm}.

%% file: sec/4_experiments.tex
\section{Experiments}
\subsection{Experimental Setup}
\noindent \textbf{Dataset.} We mainly perform experiments on the KITTI-360 dataset~\cite{liao2022kitti} for quantitative comparisons with baseline methods. KITTI-360 dataset is captured in urban scenarios covering a driving distance of around $73.7$km and offers 3D bounding box annotations for various classes (\emph{e.g.}, building, car, road, vegetation, \emph{etc}), forming extensive 3D scene layouts. Each layout instance is modeled using a triangle mesh, enabling fast rendering of semantic and depth maps. 

\noindent \textbf{Implementation Details.} We use Stable Diffusion 2.1~\cite{rombach2022high} as the text-to-image diffusion model. The overall framework is implemented using PyTorch~\cite{paszke2019pytorch}. We use diffusers~\cite{diffusers} to implement the diffusion model. AdamW~\cite{loshchilov2017decoupled} is used as the optimizer with an learning rate of $1\times 10^{-3}$. For the training of ControlNet~\cite{zhang2023adding}, we concatenate the rendered semantic and depth maps from 3D layouts as the input conditional signal. We crop and resize the original image (with a resolution of $1408\times 376$) in the KITTI-360 dataset~\cite{liao2022kitti} to a resolution of $512\times 512$. During the refinement, we further adopt monocular depth estimation method~\cite{ranftl2020towards} to predict the depth for the rendered image and align the scale and shift with the rendered depth from the generated hash grid. We then  refine the geometry by adding an L1 error between the rendered and the aligned monocular depth. Additionally, we  employ a semantic segmentation network~\cite{jain2023oneformer} to predict sky masks and encourage the accumulated density $\alpha_i$ to converge to 0 in the sky region.
The training of a single scene is conducted on a single NVIDIA V100 GPU. We initially optimize the scene at $256^2$ resolution and then refine it at $512^2$ resolution in the layout-aware refinement stage. Convergence for a scene covering a driving distance of $\sim100$m takes about $12$ hours. 

\begin{figure}[tb] % 
\centering
\includegraphics[width=1\linewidth]{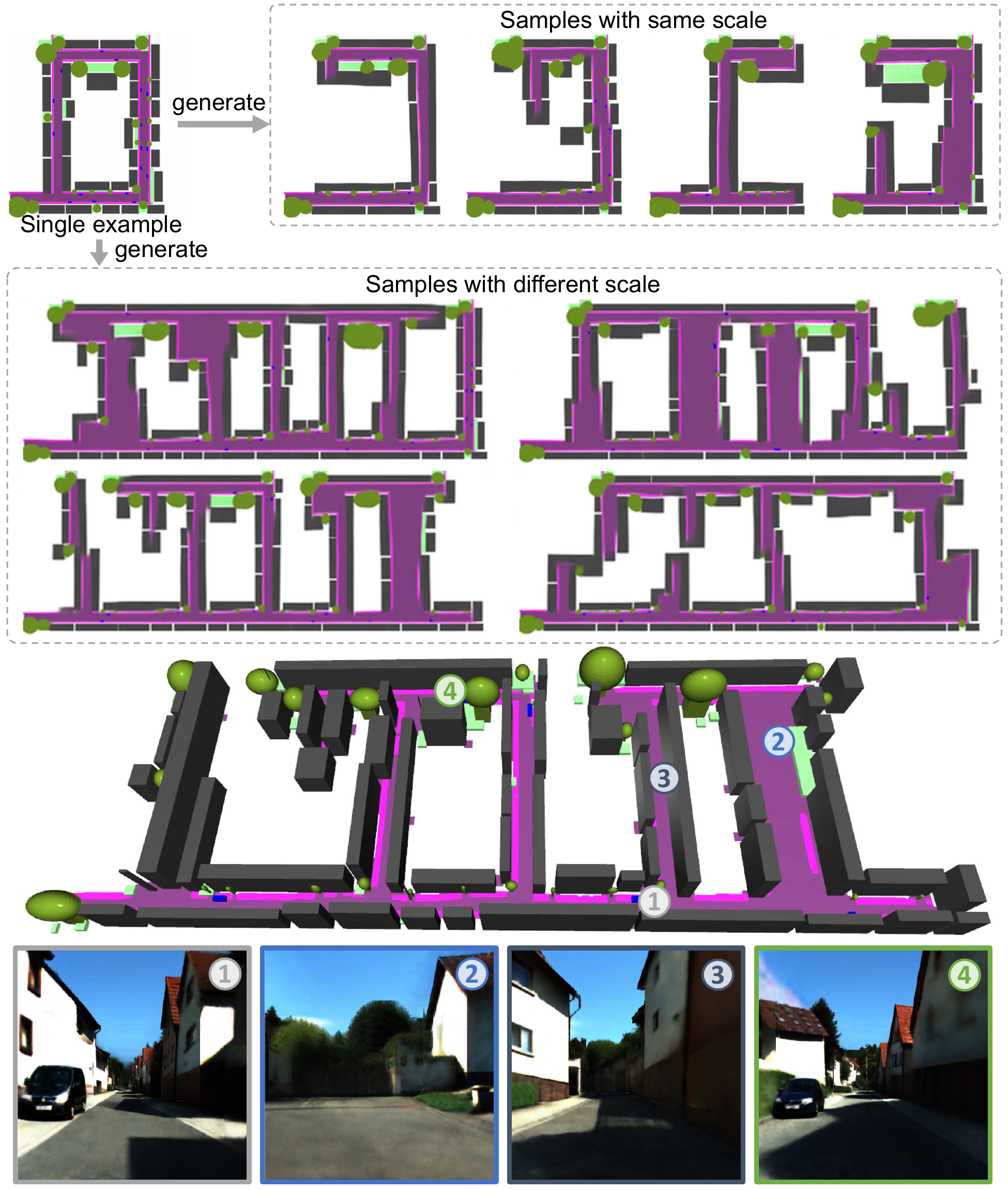}
\caption{\textbf{Automatic Layout Generation.} We present an alternative method for automatic 3D layout generation given a single example of the layout. In the top two rows, We display generated layouts with different scales and provide the corresponding 3D layout given the generated 2D sample in the third row. The rendering results of the generated scene are displayed in the bottom row.}\label{fig:sinddm}
\end{figure}

%More implementation details are provided in our supplementary materials.

% \subsection{Comparison with State-of-the-Art Methods}{\color{red}(note: where is rebutal ones?) CC3D+H}

% \begin{table}[t]
% \centering
% \setlength\tabcolsep{15pt}
% \footnotesize
% \caption{Quantitative comparison on the KITTI-360 dataset. }
% \vspace{-3ex}
% \begin{tabular}{l c c}
% \toprule
% \textbf{Method} & FID$\downarrow$ & KID$\downarrow$ \\
% \midrule
% EG3D~\cite{chan2022efficient} & 109.3 & 0.121 \\ % 121.0
% CC3D~\cite{bahmani2023cc3d} & 79.1 & 0.082 \\ % 82.3
% \midrule
% Ours & \textbf{59.8} & \textbf{0.059} \\ % 58.6
% \bottomrule
% \end{tabular}
% \label{tab:kitti}
% \vspace{-3ex}
% \end{table}

\begin{figure*}[htbp] % 
\centering
\includegraphics[width=1.\textwidth]{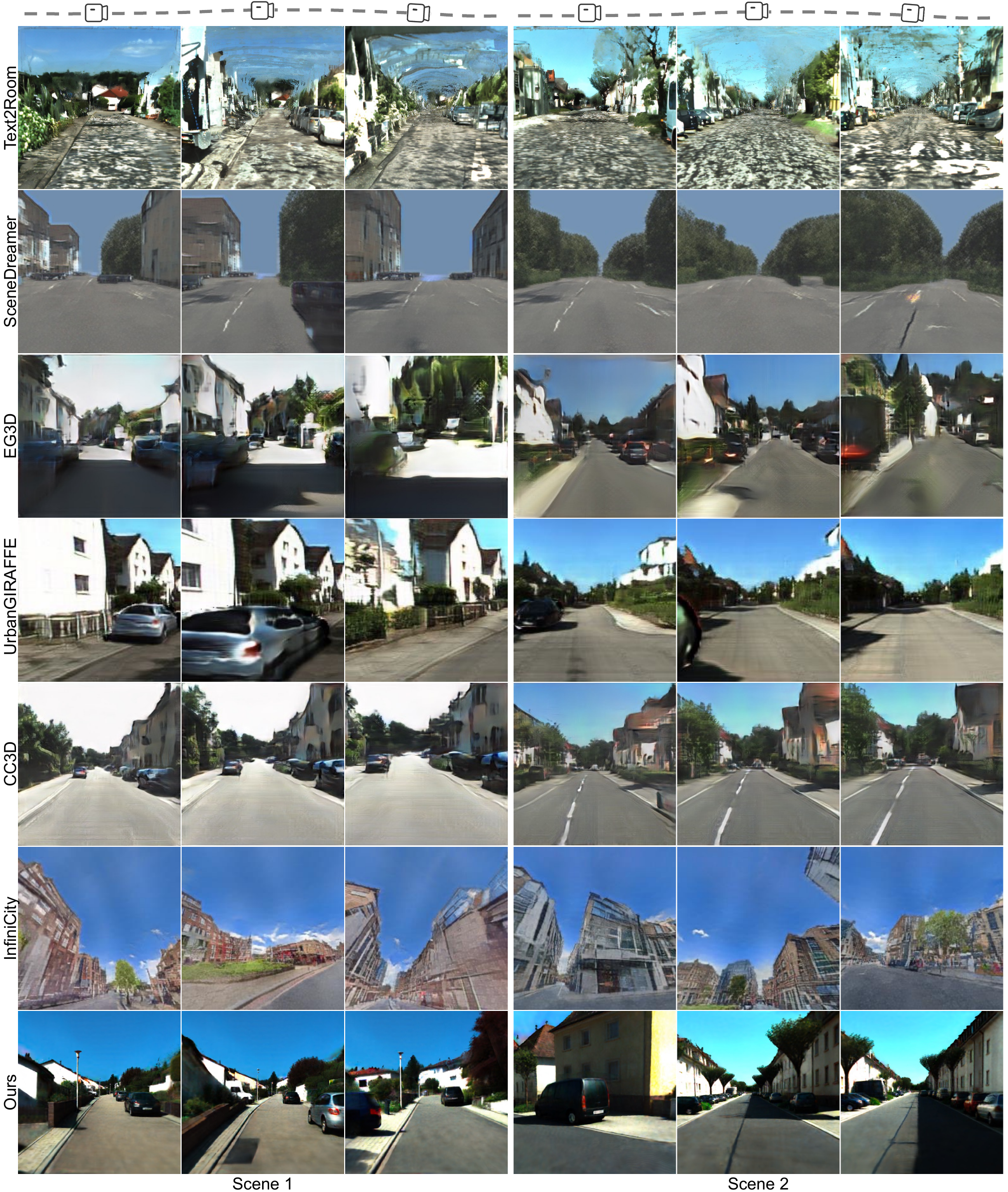}
\caption{\textbf{Qualitative comparison.} We display two scenes generated by different methods for comparison. Most of the results are directly borrowed from the original paper, except for Text2Room and SceneDreamer, which were adapted and re-trained within our settings. The proposed pipeline significantly outperforms previous baseline methods and achieves high-quality rendering. Best zoom in for more details.}\label{fig:comparison}
\end{figure*}

\begin{table*}[htbp]
\centering
\footnotesize
\caption{Quantitative comparison on the KITTI-360 dataset. }
\begin{tabular}{lcccccc}
    \toprule
        Method & Text2Room~\cite{Hollein_2023_ICCV} & UrbanGIRAFFE~\cite{yang2023urbangiraffe} & SceneDreamer~\cite{chen2023scene} & EG3D~\cite{chan2022efficient} & CC3D~\cite{bahmani2023cc3d} & Ours \\
    \midrule
        FID$\downarrow$ & $134.3$ & $118.6$ & $122.39$ & $109.3$ & $79.1$ & $\mathbf{59.8}$ \\
        KID$\downarrow$ & $0.116$ & $0.143$ & $0.113$ & $0.121$ & $0.082$ & $\mathbf{0.059}$ \\
    \bottomrule
    \end{tabular}
\label{tab:kitti}
\end{table*}

% GIRAFFE~\cite{niemeyer2021giraffe}, GSN~\cite{devries2021unconstrained}, 
\noindent\textbf{Baselines and Settings.} We compare the proposed method with several 3D generative methods applied to urban scenes, including EG3D~\cite{chan2022efficient}, CC3D~\cite{bahmani2023cc3d}, UrbanGIRAFFE~\cite{yang2023urbangiraffe}, Text2Room~\cite{Hollein_2023_ICCV}, SceneDreamer~\cite{chen2023scene} and InfiniCity~\cite{lin2023infinicity}. We use the officially released code and pre-trained weights provided by the authors of CC3D for the evaluation of CC3D and EG3D.
Due to the different settings of UrbanGIRAFFE, we re-train the model using its official released code. Text2Room is initially designed for room-scale textured mesh generation. We adapt it to our scenes by integrating our pre-trained ControlNet into the inpainting model of Text2Room, enabling the generation of urban scenes based on our 3D layout information. SceneDreamer aims to generate 3D landscapes and we apply it to urban scenes using the projected height and semantic field from our layout representation as the input. We then re-train both SceneDreamer and the required SPADE~\cite{park2019semantic} model for urban scene generation. For quantitative evaluation, we use Fréchet Inception Distance (FID)~\cite{heusel2017gans} and Kernel Inception Distance (KID)~\cite{binkowski2018demystifying} as evaluation metrics, which are computed with $5000$ randomly sampled images. Quantitative comparisons with InfiniCity are excluded due to the different settings and the incomplete released code. 

\begin{table}[tb]
\centering
\footnotesize
\caption{Quantitative comparison with CC3D+H.}
\begin{tabular}{lccc}
    \toprule
        Method & CC3D~\cite{bahmani2023cc3d} & CC3D+H~\cite{bahmani2023cc3d} & Ours \\
    \midrule
        FID$\downarrow$ & $79.1$ & $77.8$ & $\mathbf{59.8}$ \\
        KID$\downarrow$ & $0.082$ & $0.077$ & $\mathbf{0.059}$ \\
    \bottomrule
    \end{tabular}
\label{tab:cc3dh}
\end{table}

\begin{figure}[t] % 
\centering
\includegraphics[width=1\linewidth]{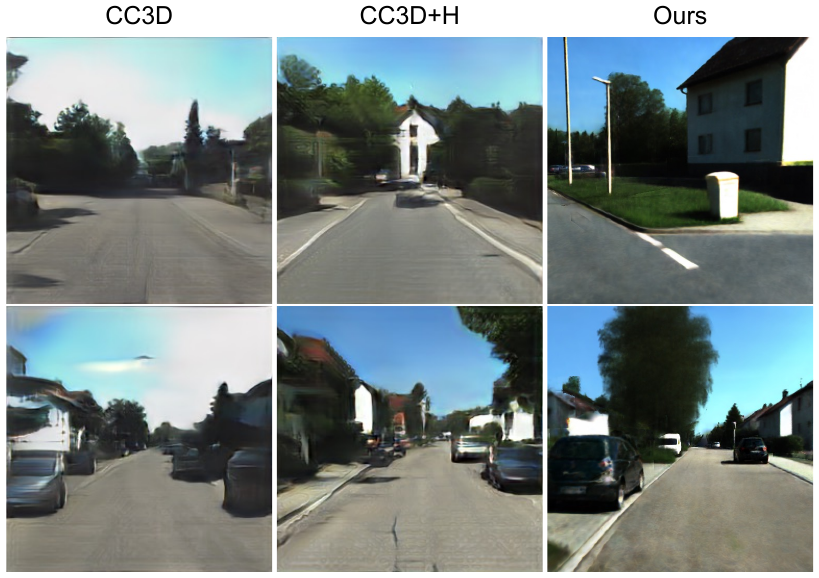}
% \vspace{-4ex}
\caption{\textbf{Qualitative comparison with CC3D+H.} Our method significantly outperforms CC3D+H, which is a enhanced version of CC3D.}\label{fig:cc3dh}
% \vspace{-4ex}
\end{figure}

\begin{figure}[t] % 
\centering
\includegraphics[width=0.85\linewidth]{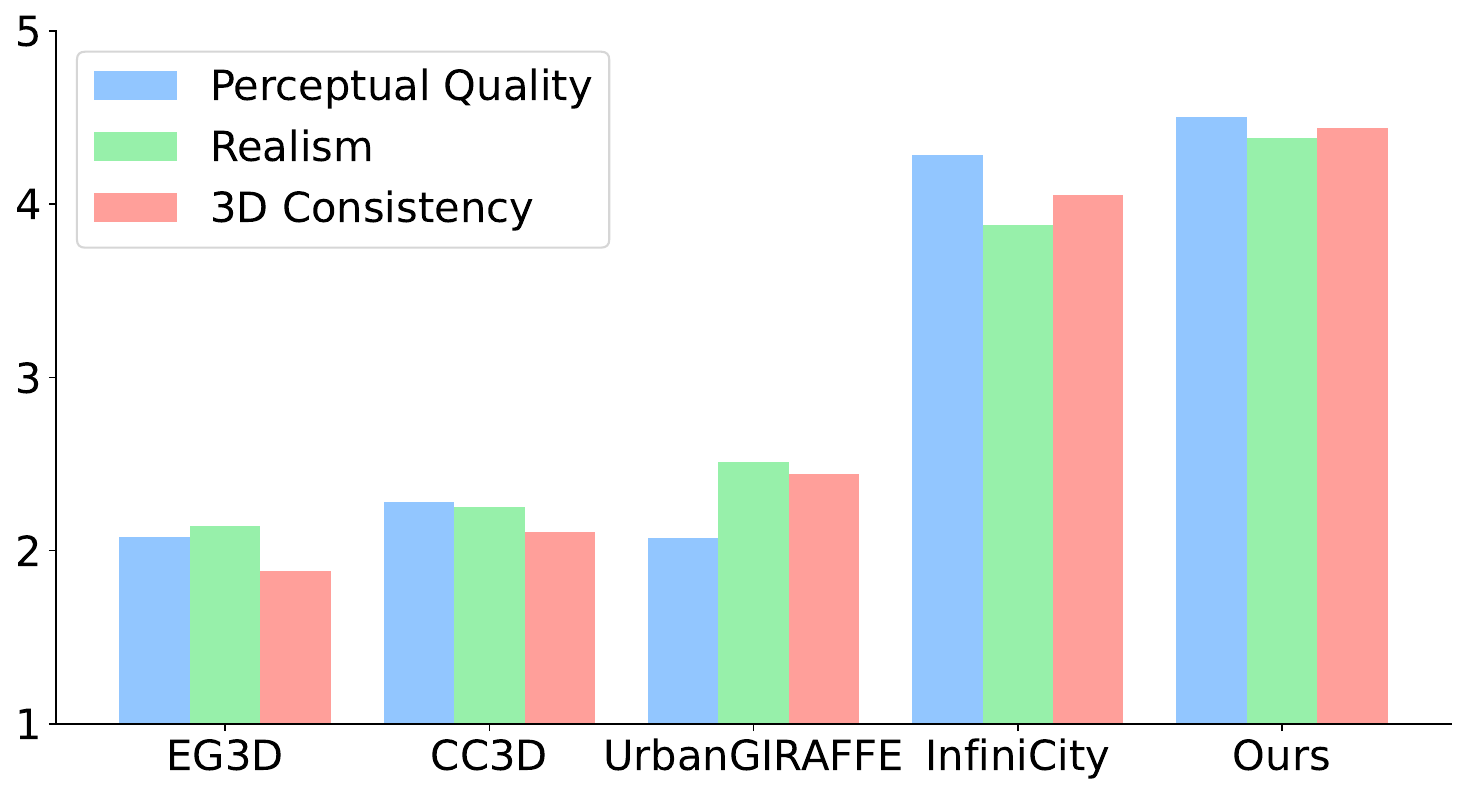}
\caption{\textbf{User study.} The score ranges from $1$ to $5$, with $5$ denoting the best quality. } \label{fig:userstudy}
\end{figure}

\begin{figure*}[tb] % 
\centering
\includegraphics[width=1\linewidth]{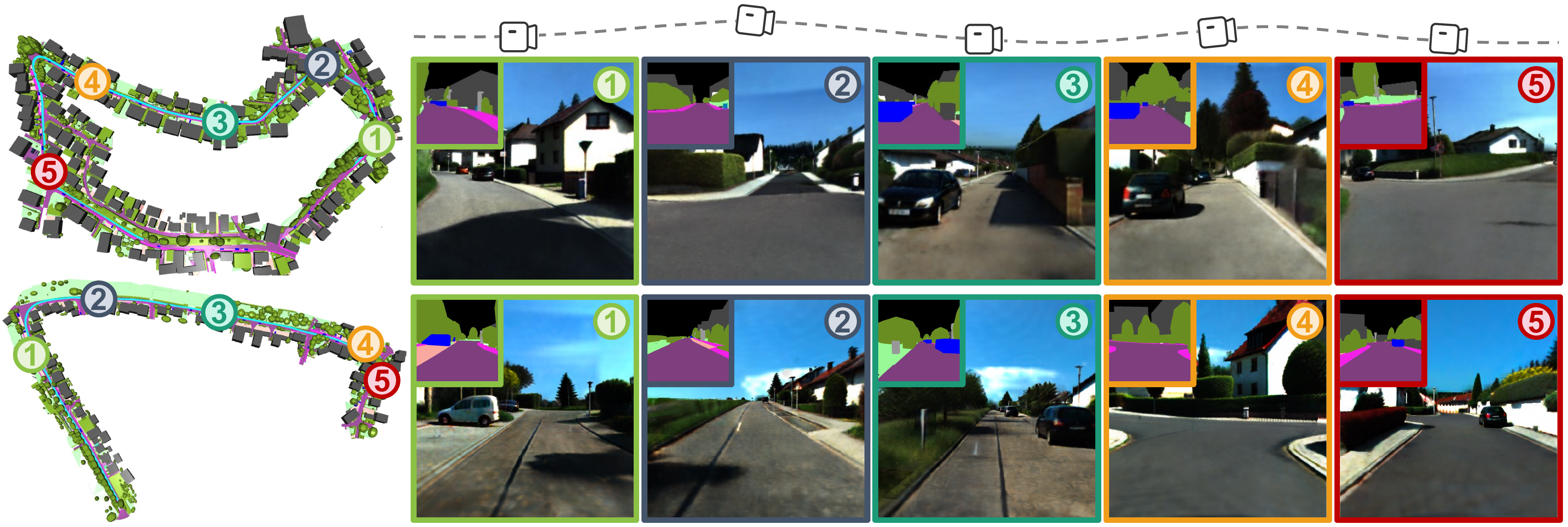}
\caption{\textbf{Large-scale generation capability.} 
We generate two large-scale 3D urban scenes with high quality, each covering an area of $\sim600\times 400 {\rm m^2}$ and spanning a driving distance of over $1000{\rm m}$.}\label{fig:largescale}
\end{figure*}

\begin{figure*}[ht] % 
\centering
\includegraphics[width=1\textwidth]{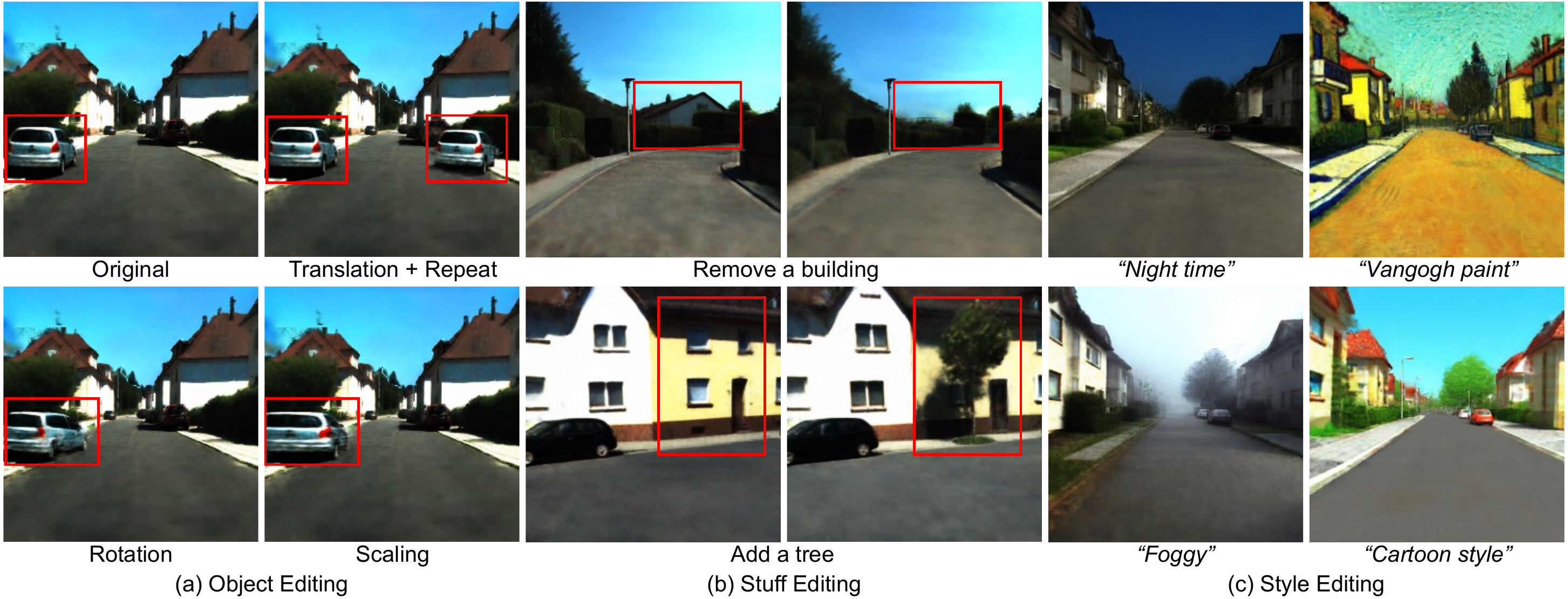}
\caption{\textbf{Samples of scene editing results.} The proposed pipeline achieves instance-level editing (including object editing and stuff editing) and global style editing.}\label{fig:editing}
\end{figure*}

\noindent\textbf{Results.} According to the quantitative results in Table~\ref{tab:kitti}, the proposed method outperforms previous methods in terms of both FID and KID by a significant margin. For example, the FID of our method is about $24\%$ lower than that of the most competitive baseline method, CC3D. We further provide visualization results of generated samples in Fig.~\ref{fig:comparison} for qualitative comparison. According to the results, Text2Room tends to degrade rapidly under the influence of cumulative errors, resulting in a failure to produce valid results.
Besides, 3D-aware GAN-based methods (\emph{i.e.}, SceneDreamer, UrbanGIRAFFE, EG3D, CC3D and Infinicity) suffer from structural distortions and lack fine-grained details, due to the limited model capacity and the complexity of urban scenes.
%The results show that EG3D fails to generate plausible scenes. Text2Room tends to degrade rapidly under the influence of cumulative errors, resulting in a failure to produce valid results. SceneDreamer XXX. UrbanGIRAFFE, CC3D, and InfiniCity show more promising results, owing to the additional conditions in the generative model. However, due to the limited model capacity and the complexity of urban scenes, they still suffer from structural distortions and lack fine-grained details. 
In contrast, our method is capable of generating high-quality 3D urban scenes with rich details, demonstrating a notable advancement over previous methods. To further investigate the effect of our 3D layout prior, we add height fields of 3D layouts as an extra condition to CC3D, resulting in an enhanced version termed CC3D+H. As shown in Table~\ref{tab:cc3dh} and Fig.~\ref{fig:cc3dh}, CC3D+H marginally outperforms CC3D, indicating that our 3D layout prior is beneficial for urban scene generation. However, it remains significantly inferior to our method, which demonstrates the effectiveness and necessity of our carefully designed pipeline to fully exploit the potential of 3D layout priors and large-scale text-to-image diffusion models.
Please refer to our project page for video results.

\noindent \textbf{User Study.} We further conduct a user study, wherein $20$ volunteers are solicited to rate the rendered videos of each method across $3$ dimensions, \emph{i.e.}, perceptual quality, realism, and 3D consistency. The score ranges from $1$ to $5$, with $5$ denoting the best quality. The results in Fig.~\ref{fig:userstudy} indicate users' preference for our results across all aspects.

\begin{figure*}[tb] % 
\centering
\includegraphics[width=0.95\textwidth]{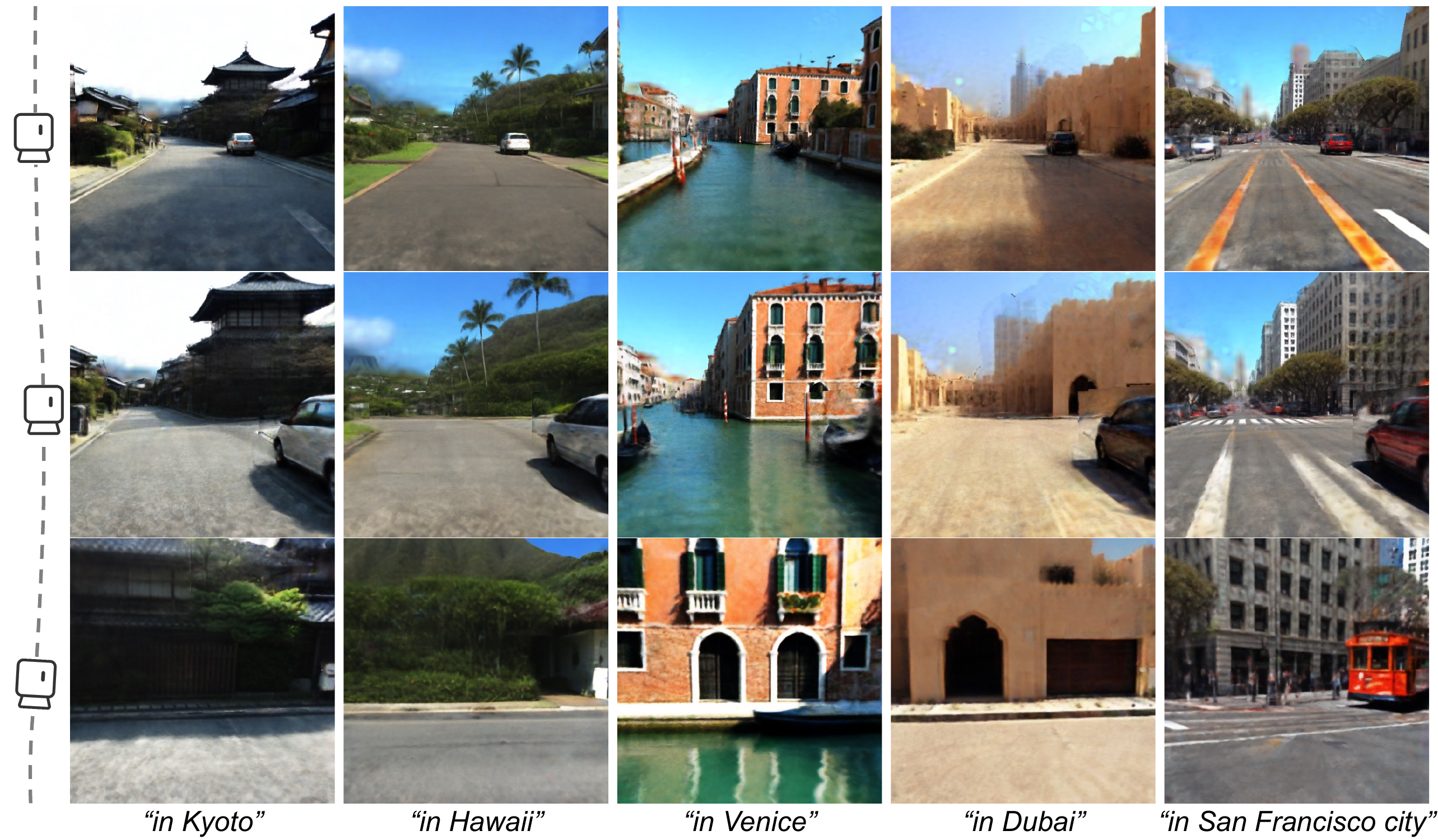}
% \vspace{-4ex}
\caption{\textbf{Transfer to other cities.} The generated urban scene can be transferred to different city styles given the text prompt by fine-tuning the generated hash grid.}
\label{fig:cityediting}
\vspace{-2ex}
\end{figure*}

% InfiniCity and SceneDreamer can generate scenes with larger scales, while the results are severely distorted.
\noindent \textbf{Large-Scale Generation Capability.} Notably, the scales of scenes generated by most prior 3D-aware GAN-based methods are limited (typically less than 100m). SceneDreamer can generate scenes with larger scales, while the resolution of a local scene window is limited to $2048\times 2048$ with 40GB GPU memory, which corresponds to a spatial coverage of $\sim 400\times 400 {\rm m^2}$ at a voxel resolution of $0.2{\rm m}$. Besides, the generated results also lack fine-grained details. In contrast, our method can achieve urban scene generation with longer driving distances and higher quality by large a margin. Fig.~\ref{fig:largescale} shows our generation results in two regions with 32GB GPU memory, each covering an area of $\sim600\times 400 {\rm m^2}$ and spanning a driving distance of over $1000{\rm m}$. Even in such large-scale scenes, our method still achieves high-quality rendering results, showcasing the superior scalability of our pipeline in the arbitrary-scale urban generation.

\subsection{Scene Editing}
\noindent \textbf{Instance-Level Editing.} As shown in Fig.~\ref{fig:editing} (a) (b), we showcase two kinds of instance-level editing: \textit{(a) Object manipulation.} We achieve object rotation, translation, repeat, and scaling by simply manipulating the object in the layout. \textit{(b) Stuff editing.} We can further edit stuff elements in the scene (\emph{e.g.}, building removing, tree adding) by editing the layouts while keeping the other components unchanged.

\noindent \textbf{Style Editing.} We present several style editing samples corresponding to the same generated scene in Fig.~\ref{fig:editing} (c). The results show that we can transfer the style of the generated scene via style text prompts (\emph{e.g.}, \textit{``Foggy"}, \textit{``Vangogh paint"}, \emph{etc}), while preserving the overall structure. Notably, we can also generate urban scene of various city styles by simply adding text prompt-based conditions (\emph{e.g.}, \textit{``in Kyoto"}, \textit{``in Hawaii"}, \emph{etc}).

\begin{figure}[t] % 
\centering
\includegraphics[width=1\linewidth]{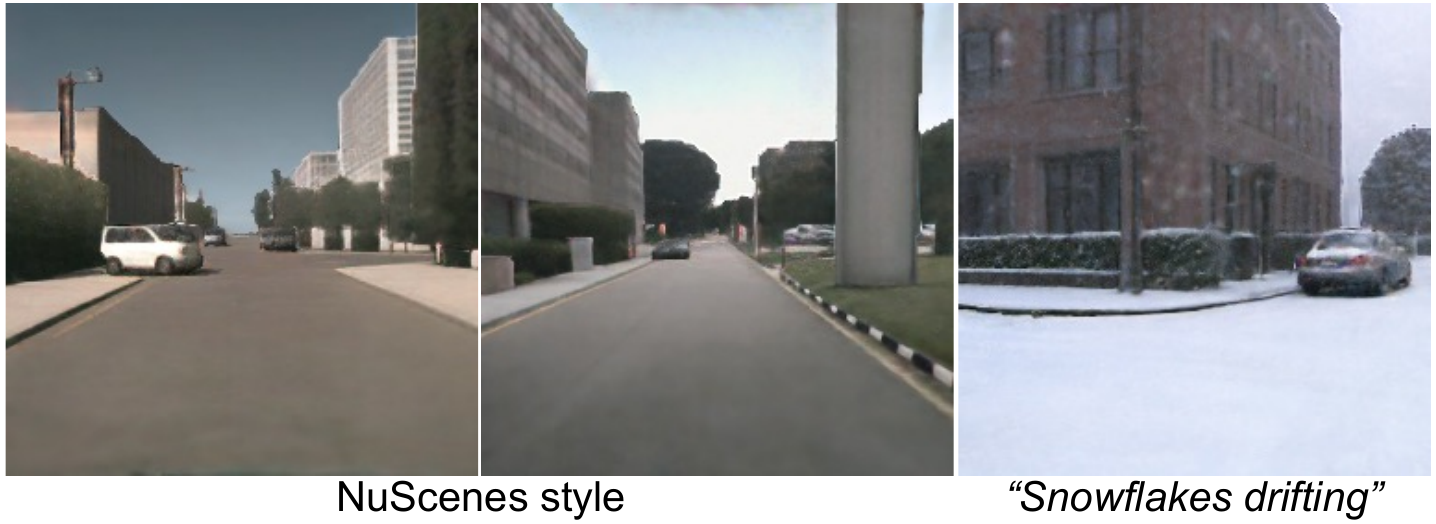}
\caption{\textbf{Transfer to other urban styles.} We generate scenes in NuScenes style and transfer them to a different weather.}\label{fig:transfer}
\end{figure}

\noindent \textbf{Transfer to Other Scenes.} Although the conditioning model is initially trained on the KITTI-360 dataset, we can transfer our method to generate scenes of other urban datasets by fine-tuning the ControlNet. As shown in Fig.~\ref{fig:transfer}, we generate scenes on the NuScenes~\cite{caesar2020nuscenes} dataset and further transfer them into a different style (\emph{i.e.}, \textit{``Snowflakes drifting"}).

\subsection{Ablation Studies}

\begin{figure}[t] % 
\centering
\includegraphics[width=1.\linewidth]{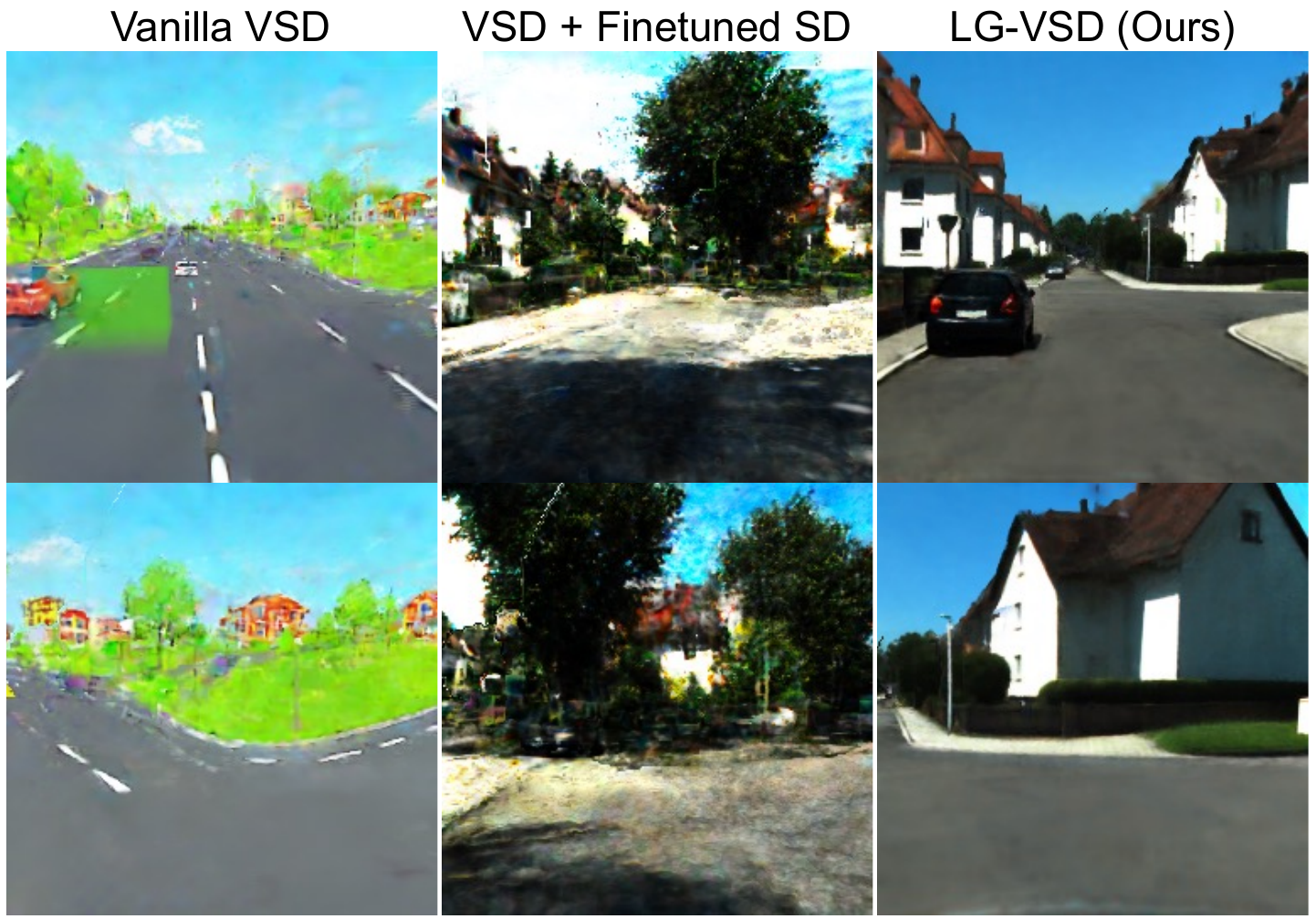}
\caption{\textbf{Effectiveness of LG-VSD.} Without our 3D layout prior, VSD fails to generate high-quality urban scenes due to the absence of effective guidance, even with the fine-tuned diffusion model.}\label{fig:lg-vsd}
\end{figure}

\noindent \textbf{Effectiveness of LG-VSD.} To verify the effectiveness of LG-VSD, we first conduct experiments by directly using the vanilla VSD, which does not yield reasonable results. By further applying our layout-constrained sampling strategy, we achieved more plausible results (shown in the left column of Fig.~\ref{fig:lg-vsd}). Nonetheless, due to the inherent ambiguity of text prompts, VSD tends to converge to a trivial solution, resulting in unrealistic results that lack fine-grained details. To obtain realistic urban-like generation results, we further fine-tune the noise prediction network of Stable Diffusion on the KITTI-360 dataset to constrain the style distribution. However, in the absence of semantic and geometric constraints from the 3D layout, VSD fails to capture the complex distribution of urban scenes, yielding unsatisfactory results (shown in the middle column of Fig.~\ref{fig:lg-vsd}). Quantitative comparison in Table~\ref{tab:ablation} also indicates that our LG-VSD exhibits much lower FID and KID than VSD.

\begin{table}[tb]
\centering
\renewcommand\arraystretch{1.0}
\setlength\tabcolsep{10pt}
\footnotesize
\caption{Ablation studies on LG-VSD and layout-constrained sampling strategy on the KITTI-360 dataset.}
% KID is multiplied by $1000\times$ according to CC3D.
% \vspace{-3ex}
\begin{tabular}{l c c}
\toprule
\textbf{Method} & FID$\downarrow$ & KID$\downarrow$ \\
\midrule
w/o LG-VSD & 167.1 & 0.203 \\ % 121.0
w/o layout-constrained sampling & 143.5 & 0.148 \\ % 82.3
\midrule
Ours & \textbf{59.8} & \textbf{0.059} \\ % 58.6
\bottomrule
\end{tabular}
\label{tab:ablation}
% \vspace{-3ex}
\end{table}

\begin{figure}[tb] % 
\centering
\includegraphics[width=1.\linewidth]{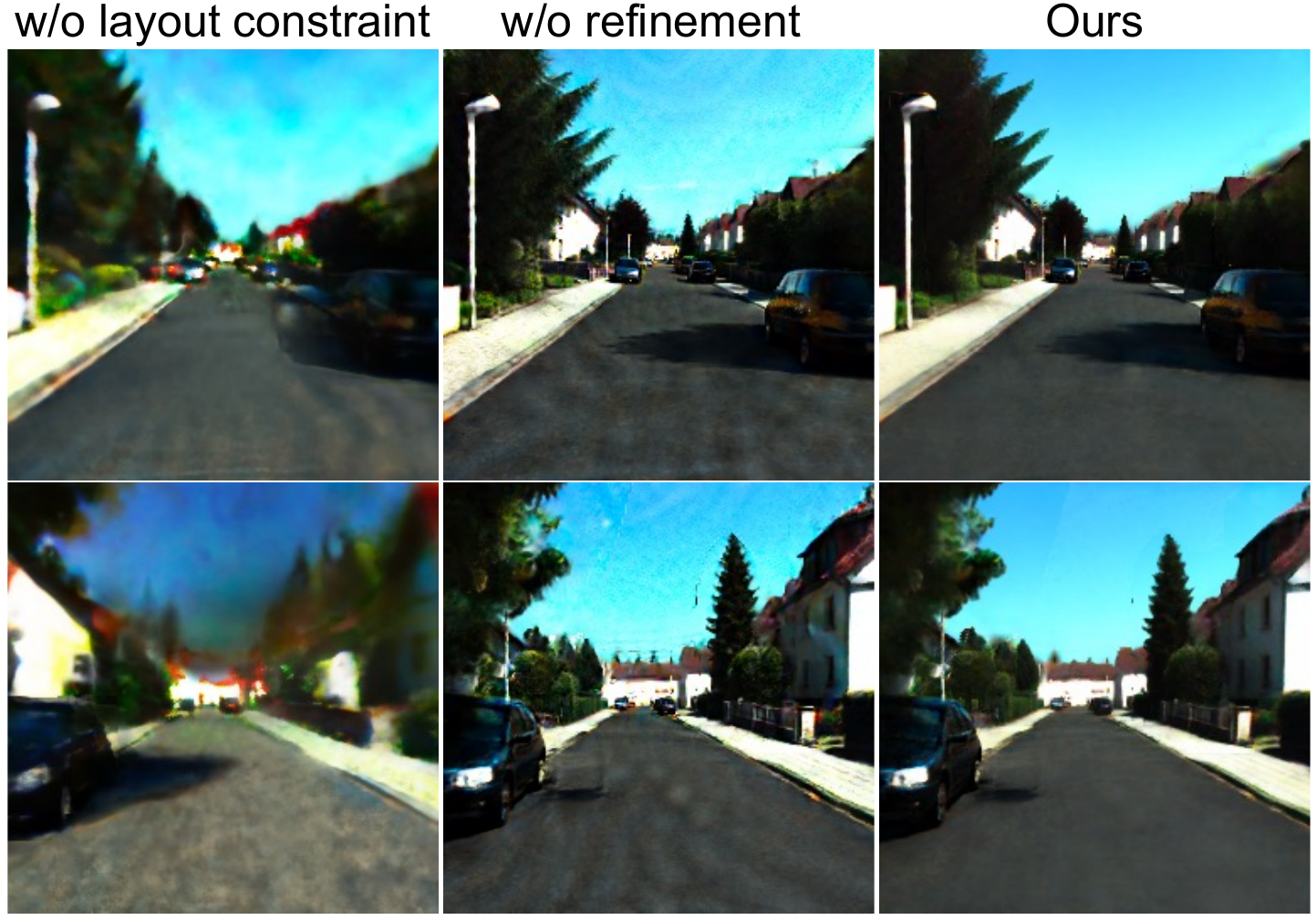}
\caption{\textbf{Ablations on layout-constrained sampling and layout-aware refinement strategies.} The rendering results are blurry without the layout-constrained sampling strategy. The layout-aware refinement strategy further enhances the generation quality, leading to more realistic results.}\label{fig:refinemnet}
\end{figure}

\noindent \textbf{Layout-Constrained Sampling.} To investigate the impact of the layout-constrained sampling strategy (\emph{i.e.}, constraining the sampling space of our scalable hash grid using the 3D layout prior), we perform experiments by dropping this constraint (\emph{i.e.}, sampling in the full space). As shown in the left column of Fig.~\ref{fig:refinemnet}, the proposed pipeline can still produce plausible results without the constraint, while yielding much blurrier rendering results. Results in Table~\ref{tab:ablation} further demonstrate the effectiveness of our layout-constrained sampling strategy.

\noindent \textbf{Layout-Aware Refinement.} We perform experiments to analyze the effectiveness of the layout-aware refinement strategy. As shown in the middle column of Fig.~\ref{fig:refinemnet}, aliasing artifacts are evident without the refinement, while the refined results are smoother and more realistic.

% \subsection{Results}

\begin{figure}[tb] % 
\centering
\includegraphics[width=1\linewidth]{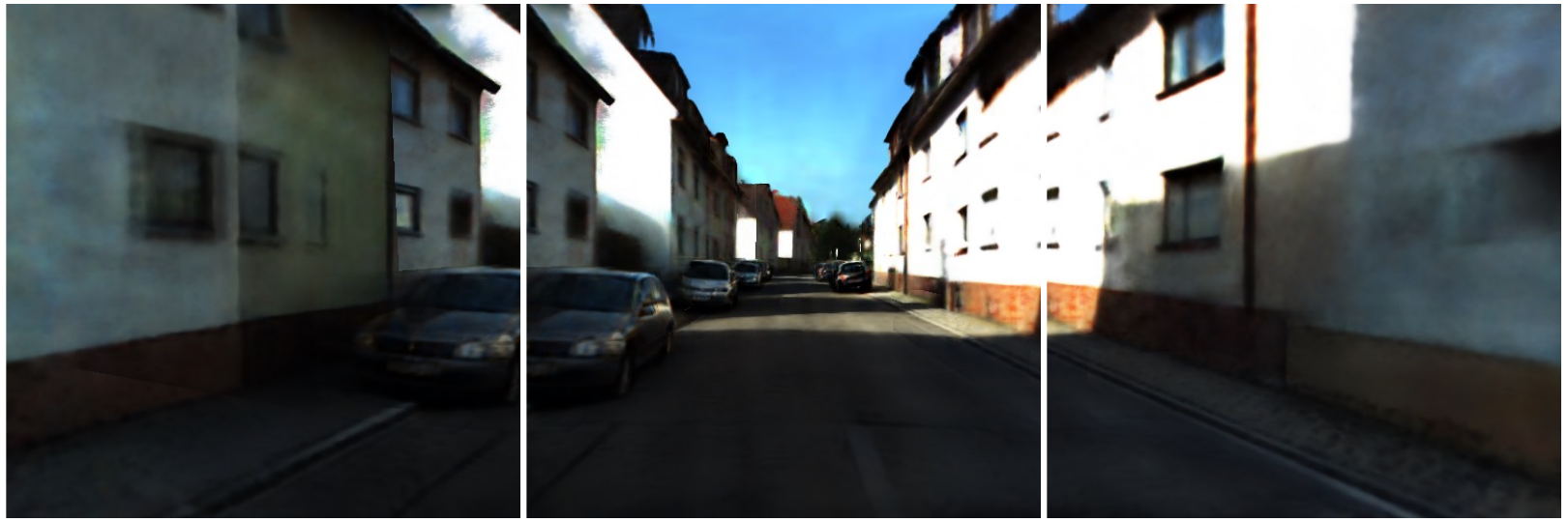}
\caption{\textbf{Large camera view shifting.} The camera rotates from $-45\deg$ to $45\deg$ from left to right.}\label{fig:viewshifting}
\end{figure}

\begin{figure}[ht] % 
\centering
\includegraphics[width=1\linewidth]{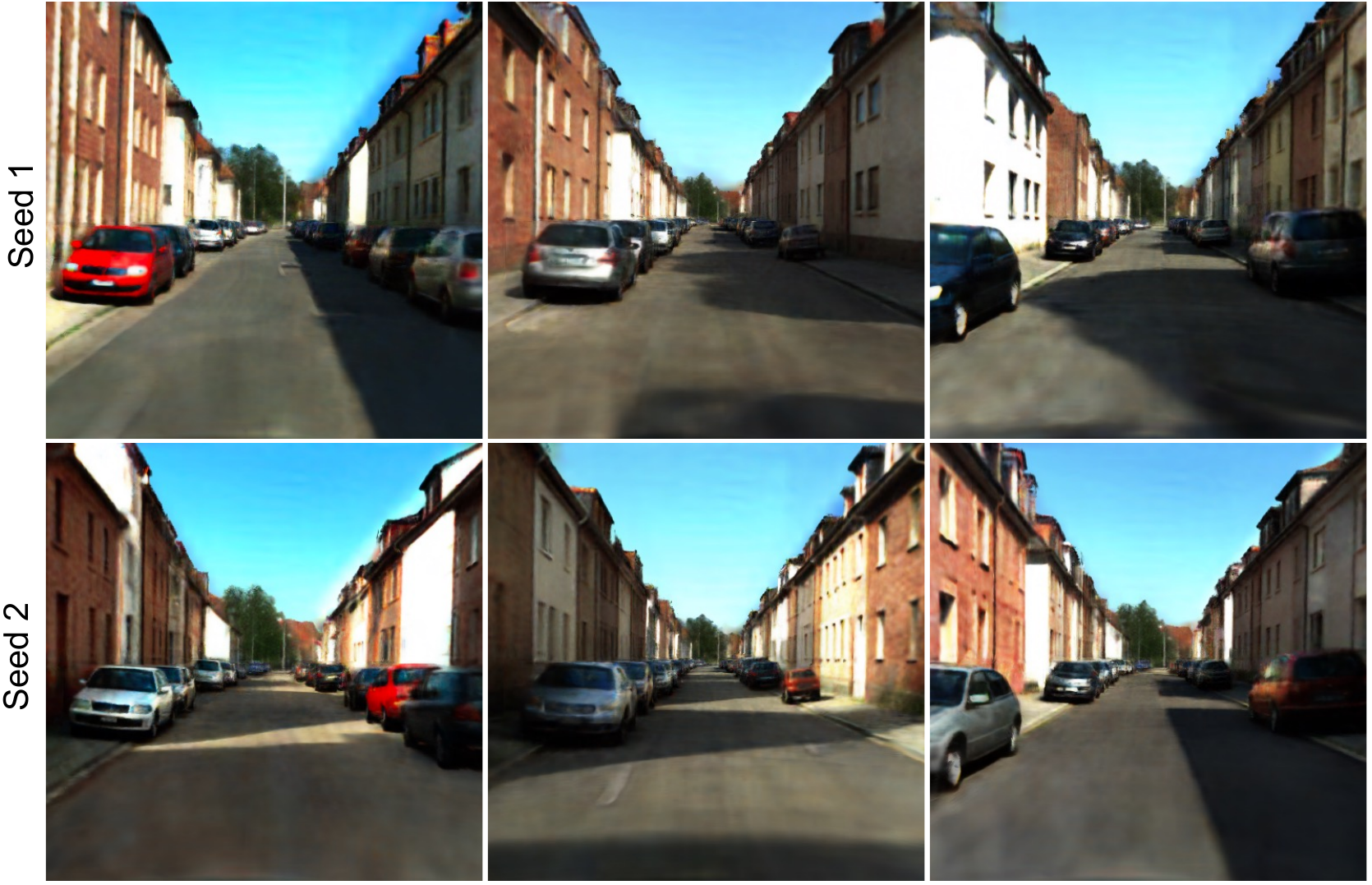}
\caption{\textbf{Generated results with different random seeds.} We display the generated scenes with two different random seeds given the same scene layout.}\label{fig:diversity}
\end{figure}

\begin{figure}[tb] % 
\centering
\includegraphics[width=1\linewidth]{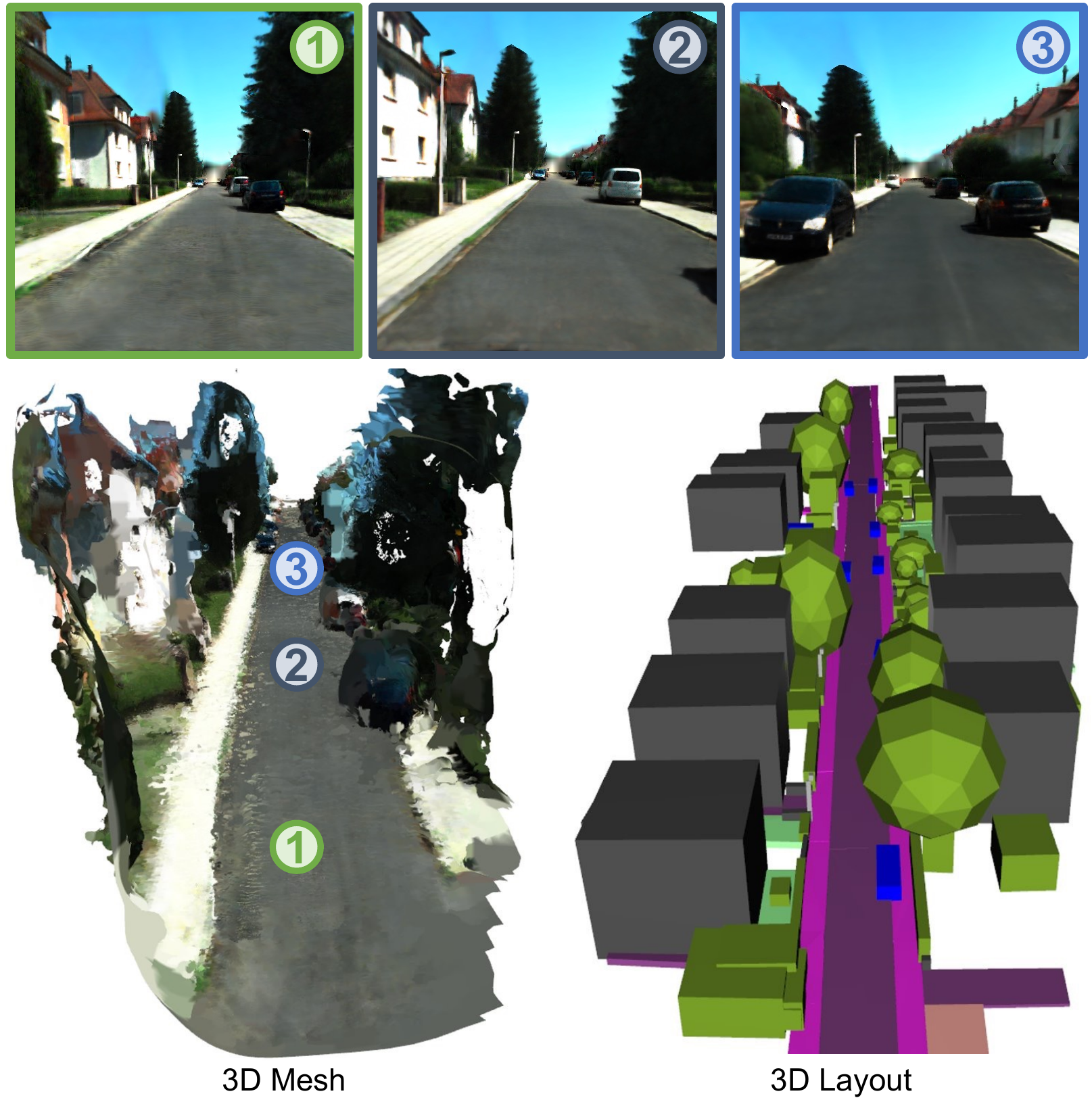}
\caption{\textbf{3D mesh visualization.} We extract a 3D triangle mesh from the generated scene and provide rendered 2D images from corresponding camera poses.}
\label{fig:3dresults}
\end{figure}

\subsection{Further Analysis}

\noindent\textbf{Large Camera View Shifting.} As shown in Fig.~\ref{fig:viewshifting}, we display the rendering results by rotating the camera from $-45\deg$ to $45\deg$ from the training trajectory. Owing to the high 3D consistency, the generated scene demonstrates strong robustness against large camera view shifting.

\noindent\textbf{Diversity.} To explore the diversity of generated results, we conduct experiments by employing different random seeds while maintaining the same layout. As shown in Fig.~\ref{fig:diversity}, given the same scene layout, the proposed pipeline can generate diverse scenes with elements (\emph{e.g.}, cars, buildings) that have different appearances and illumination.

\noindent\textbf{3D Visualization.} To further explore the 3D consistency of the generated scene, we extract a triangle mesh from the generated hash grid representation and display the results in Fig.~\ref{fig:3dresults}. The 3D triangle mesh reveals the consistent 3D structures of the generated scene.

%% file: sec/5_conclusion.tex
\section{Conclusion and Limitations}
We have presented Urban Architect, a method for steerable 3D urban scene generation. The core of our method lies in incorporating a 3D layout representation as a robust prior to complement textual descriptions. Based on the 3D layout, we introduce Layout-Guided Variational Score Distillation (LG-VSD) to integrate the 3D geometric and semantic constraints into the current text-to-3D paradigm, which is achieved by conditioning the score distillation sampling process with 3D layout information. Moreover, to address the unbounded nature of urban scenes, we design a scalable hash grid representation to adapt to arbitrary scene scales. Our framework facilitates high-quality, steerable 3D urban scene generation, capable of scaling up to generate large-scale scenes covering a driving distance of over 1000m. Our flexible representation, coupled with the inherent capability of text-to-image diffusion models, empowers our pipeline to support diverse scene editing effects, including instance-level editing and style editing. Despite achieving high-quality and steerable generation, the current optimization process cannot satisfy pixel-level scene control. We hope our method can be a starting point for the urban-scale scene generation task, and leave addressing the above issues as future work.